\documentclass{article}

    \usepackage[final, nonatbib]{neurips_2022}

\usepackage[utf8]{inputenc} % allow utf-8 input
\usepackage[T1]{fontenc}    % use 8-bit T1 fonts
\usepackage{hyperref}       % hyperlinks
\usepackage{url}            % simple URL typesetting
\usepackage{booktabs}       % professional-quality tables
\usepackage{amsfonts}       % blackboard math symbols
\usepackage{nicefrac}       % compact symbols for 1/2, etc.
\usepackage{microtype}      % microtypography
\usepackage{amsmath}
\usepackage{graphicx}
\usepackage{colortbl}
\usepackage{xcolor}
\usepackage{wrapfig}
\usepackage{subcaption}
\usepackage{multirow}
\usepackage{makecell}
\usepackage{mathrsfs}
\usepackage{algorithm}
\usepackage{algpseudocode}
\usepackage{enumitem}
\usepackage{multirow}

\usepackage{listings}
\lstdefinestyle{mystyle}{
    commentstyle=\color{codegreen},
    keywordstyle=\color{magenta},
    numberstyle=\tiny\color{codegray},
    stringstyle=\color{codepurple},
    basicstyle=\ttfamily\footnotesize,
    breakatwhitespace=false,         
    breaklines=true,                 
    captionpos=b,                    
    keepspaces=true,                 
    numbers=left,                    
    numbersep=5pt,                  
    showspaces=false,                
    showstringspaces=false,
    showtabs=false,                  
    tabsize=2
}

\definecolor{codegreen}{rgb}{0,0.6,0}
\definecolor{codegray}{rgb}{0.5,0.5,0.5}
\definecolor{codepurple}{rgb}{0.58,0,0.82}
\definecolor{backcolour4Input}{rgb}{0.85,0.95,1}
\definecolor{backcolour4Output}{rgb}{1,0.85,0.95}

\title{\raggedright PRIME-CVD:\\
A Parametrically Rendered Informatics Medical Environment\\
for Education in Cardiovascular Risk Modelling}

\author{Nicholas I-Hsien Kuo,\\ \textbf{Marzia Hoque Tania,}\\
\textbf{Blanca Gallego, Louisa R Jorm}\\
Centre for Big Data Research in Health (CBDRH)\\
The University of New South Wales, Sydney, Australia\\
\footnotesize{\textcolor{white}{*}}\\
Corresponding author: Nicholas I-Hsien Kuo (\texttt{n.kuo@unsw.edu.au})
}

\begin{document}

\maketitle

\begin{abstract}
    In recent years, progress in medical informatics and machine learning has been accelerated by the availability of openly accessible benchmark datasets. However, patient-level electronic medical record (EMR) data are rarely available for teaching or methodological development due to privacy, governance, and re-identification risks. This has limited reproducibility, transparency, and hands-on training in cardiovascular risk modelling. Here we introduce PRIME-CVD, a parametrically rendered informatics medical environment designed explicitly for medical education. PRIME-CVD comprises two openly accessible synthetic data assets representing a cohort of 50,000 adults undergoing primary prevention for cardiovascular disease. The datasets are generated entirely from a user-specified causal directed acyclic graph parameterised using publicly available Australian population statistics and published epidemiologic effect estimates, rather than from patient-level EMR data or trained generative models. Data Asset 1 provides a clean, analysis-ready cohort suitable for exploratory analysis, stratification, and survival modelling, while Data Asset 2 restructures the same cohort into a relational, EMR-style database with realistic structural and lexical heterogeneity. Together, these assets enable instruction in data cleaning, harmonisation, causal reasoning, and policy-relevant risk modelling without exposing sensitive information. Because all individuals and events are generated de novo, PRIME-CVD preserves realistic subgroup imbalance and risk gradients while ensuring negligible disclosure risk. PRIME-CVD is released under a Creative Commons Attribution 4.0 licence to support reproducible research and scalable medical education.
\end{abstract}

%###===>>>%###===>>>%###===>>>
%###===>>>%###===>>>%###===>>>
%###===>>>%###===>>>%###===>>>
\newpage
\section{Background \& Summary}
%%%%%%%%%%%%%%%%%%%%%%%%%%%%%%%%%%%%%%%%%%%%%%%%%%%%%%%%%%%%
%%%%%%%%%%%%%%%%%%%%%%%%%%%%%%%%%%%%%%%%%%%%%%%%%%%%%%%%%%%%
%%%%%%%%%%%%%%%%%%%%%%%%%%%%%%%%%%%%%%%%%%%%%%%%%%%%%%%%%%%%
\subsection{Introduction}
Medical informatics -- first articulated by Fran\c{c}ois Gr\'emy as the intersection of clinical practice, statistical reasoning, and computational methods~\cite{degoulet2005francois} -- now underpins every layer of modern healthcare delivery. Medical informaticians ensure that health information is accurate, interoperable, and actionable; they develop risk-prediction models~\cite{nicholas2025estimating}, optimise hospital operations~\cite{shawon2024non}, support system-wide decision-making~\cite{peiris2024overcoming}, and evaluate emerging treatments across the cardio-kidney-metabolic (CKM) spectrum~\cite{wallace2025prevalence}. Preparing the next generation of informaticians therefore requires a curriculum that integrates clinical domain knowledge with epidemiology, causal inference, data engineering, and modern computational tools. In practice, however, educational programs often struggle to provide authentic, hands-on experience with electronic medical record (EMR) data. Real patient-level data are difficult to access, tightly governed, and frequently withheld from teaching due to privacy concerns. Even when de-identified resources are available, they require substantial cleaning and harmonisation expertise~\cite{liu2022identifying} that is challenging to teach safely using real patients.

Hands-on learning frameworks consistently show that students learn best by working directly with realistic datasets~\cite{prince2004does, nicholas2024enriching}. Yet in health informatics, the absence of openly accessible, privacy-preserving, EMR-like data remains a fundamental barrier to effective training.

%%%%%%%%%%%%%%%%%%%%%%%%%%%%%%%%%%%%%%%%%%%%%%%%%%%%%%%%%%%%
%%%%%%%%%%%%%%%%%%%%%%%%%%%%%%%%%%%%%%%%%%%%%%%%%%%%%%%%%%%%
%%%%%%%%%%%%%%%%%%%%%%%%%%%%%%%%%%%%%%%%%%%%%%%%%%%%%%%%%%%%
\subsection{Related Work}
Several openly accessible health datasets have transformed research and training, most notably the MIMIC-series of critical-care databases~\cite{johnson2023mimic}, which provide rich, de-identified EMR records for methodological development and benchmarking. However, because MIMIC and similar EMR resources require credentialled access, each student must clear the same administrative hurdle, creating a bottleneck that makes it difficult to design courses in which large cohorts can use the data for assignments or examinations. Synthetic datasets have therefore been explored as substitutes for real EMR, such as the synthetic Cancer Registry Data produced by Simulacrum for NHS England~\cite{kafatos2025leveraging}. Many recent synthetic EMR resources rely on neural network models -- including generative adversarial networks (GANs)~\cite{goodfellow2014generative, kuo2022health}, denoising diffusion probabilistic models (DDPMs)~\cite{ho2020denoising, nicholas2023synthetic}, and autoregressive Transformer-based approaches~\cite{vaswani2017attention, theodorou2023synthesize, kuo2025limits}. Although these methods show strong generative performance, they learn directly from real patient trajectories and thus retain residual membership-inference risks~\cite{el2020evaluating}, limiting their suitability for education. 

To address this, we introduce PRIME -- the Parametrically Rendered Informatics Medical Environment. PRIME provides a fully transparent, pedagogically oriented cohort of simulated patients for cardiovascular disease (CVD) prognosis, generated from a directed acyclic graph (DAG)~\cite{lipsky2022causal} parameterised exclusively using openly available Australian government statistics (\textit{e.g.,} AIHW) and published literature.

%%%%%%%%%%%%%%%%%%%%%%%%%%%%%%%%%%%%%%%%%%%%%%%%%%%%%%%%%%%%
%%%%%%%%%%%%%%%%%%%%%%%%%%%%%%%%%%%%%%%%%%%%%%%%%%%%%%%%%%%%
%%%%%%%%%%%%%%%%%%%%%%%%%%%%%%%%%%%%%%%%%%%%%%%%%%%%%%%%%%%%
\subsection{Data Asset Characteristics}

\begin{figure}[h!]
    \centering
    \includegraphics[width=\linewidth]{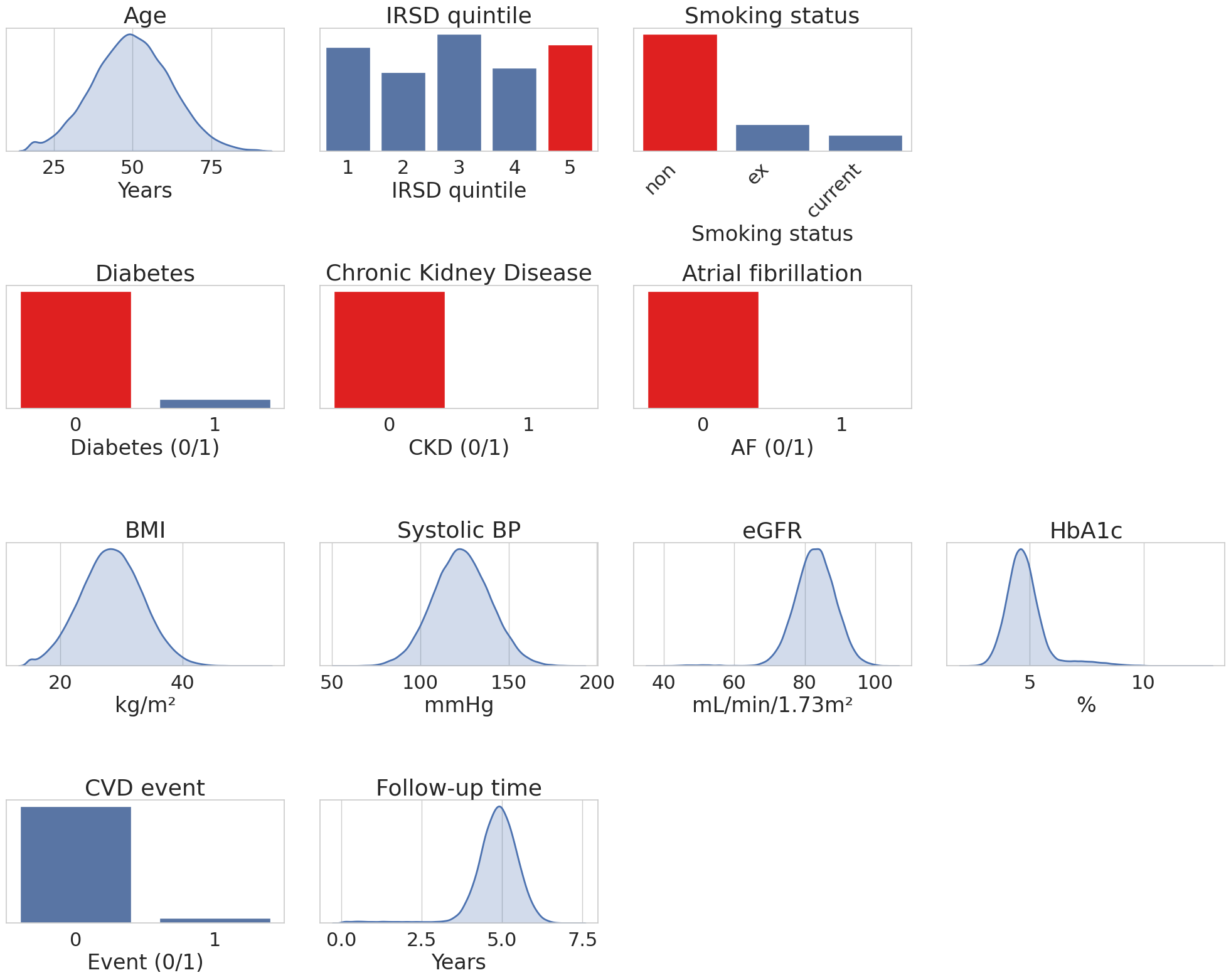}
    \caption{Baseline characteristics of the PRIME-CVD cohort used for cardiovascular risk simulation.}
    \label{fig:prime_baseline_dist}
\end{figure}

\textbf{Data Asset 1}\\
PRIME-CVD Data Asset~1 comprises 50{,}000 simulated adults aged 18--90 years
(median 49.6; Q1--Q3: 41.3--58.1). IRSD quintiles are approximately balanced
across the cohort, with around 20\% of individuals in each deprivation stratum.
At baseline, 73.1\% are non-smokers, 10.1\% current smokers, and 16.7\% ex-smokers.
The prevalences of diabetes, chronic kidney disease, and atrial fibrillation are
7.4\%, 0.7\%, and 0.7\%, respectively. Median BMI is 28.3\,kg/m$^2$
(Q1--Q3: 24.9--31.7), median systolic blood pressure is 123.1\,mmHg
(Q1--Q3: 112.4--134.1), and median eGFR is 82.9\,mL/min/1.73m$^2$
(Q1--Q3: 79.2--86.7). Over a nominal 5-year follow-up, the composite CVD outcome
occurs in approximately 4\% of individuals, with mean follow-up 4.8 years.
These characteristics are summarised in Table~\ref{Table:Table01} and
distributional visualisations in Figure~\ref{fig:prime_baseline_dist}.

\textbf{Data Asset 2}\\
Data Asset~2 restructures the same 50{,}000 adults into a relational, EMR-style database with three linked tables (Table~\ref{Tab:Dic}). \texttt{PatientMasterSummary} contains identifiers, demographics, socioeconomic position, smoking status, and a coarsened five-year CVD outcome. \texttt{PatientChronicDiseases} expands baseline conditions into heterogeneous free-text and code-like labels with diffuse diagnosis dates. \texttt{PatientMeasAndPath} stores anthropometric measurements in long form with mixed units and varied description fields. Together, these tables reproduce the structural and lexical messiness of real primary-care EMR and require linkage and cleaning to recover the underlying cohort.

%###===>>>%###===>>>%###===>>>
%###===>>>%###===>>>%###===>>>
%###===>>>%###===>>>%###===>>>
%\newpage
\section{Methods}

PRIME-CVD comprises two datasets designed for medical education and methodological training:

\hspace*{5mm}
\begin{minipage}{0.9\textwidth}
\textbf{Data Asset 1} is a fully specified, epidemiologically coherent cohort generated using a parametric, DAG-based simulation engine. This dataset represents an idealised\newline
population in which all variables follow known causal mechanisms and exhibit internally consistent distributions and risk relationships.
\end{minipage}

\hspace*{5mm}
\begin{minipage}{0.9\textwidth}
\textbf{Data Asset 2} is a systematic transformation of Data Asset 1 into a multi-table, EMR-style data resource that captures the structural and lexical heterogeneity characteristic of routinely collected clinical records. The dataset introduces realistic challenges such as inconsistent coding systems, heterogeneous measurement units, variable naming inconsistency, and cross-table linkage requirements.
\end{minipage}

This section describes the construction of both datasets.

%%%%%%%%%%%%%%%%%%%%%%%%%%%%%%%%%%%%%%%%%%%%%%%%%%%%%%%%%%%%
%%%%%%%%%%%%%%%%%%%%%%%%%%%%%%%%%%%%%%%%%%%%%%%%%%%%%%%%%%%%
%%%%%%%%%%%%%%%%%%%%%%%%%%%%%%%%%%%%%%%%%%%%%%%%%%%%%%%%%%%%
\subsection{Data Asset 1: A Clean Cohort Assembled from a Fully Parameterised Graph}
\textbf{Data Sources, Parametrisation, and Cohort Assembly}\\
The assembly engine is parameterised using publicly accessible data sources, including population demographic distributions from the Australian Bureau of Statistics (ABS), chronic-disease prevalence summaries from the Australian Institute of Health and Welfare (AIHW), and effect estimates from large cohort studies of cardiometabolic and renal risk. These inputs define the target distributions for socioeconomic status (IRSD quintiles) and age in primary-prevention populations; the marginal prevalences of diabetes, chronic kidney disease (CKD), and atrial fibrillation (AF); and the population-level means and variances of BMI, systolic blood pressure (SBP), eGFR, and HbA1c. Published associations between demographic, lifestyle, anthropometric, and clinical factors further guide the specification of directional relationships and their approximate magnitudes.

\begin{figure}[h!]
    \centering
    \includegraphics[width=0.75\linewidth]{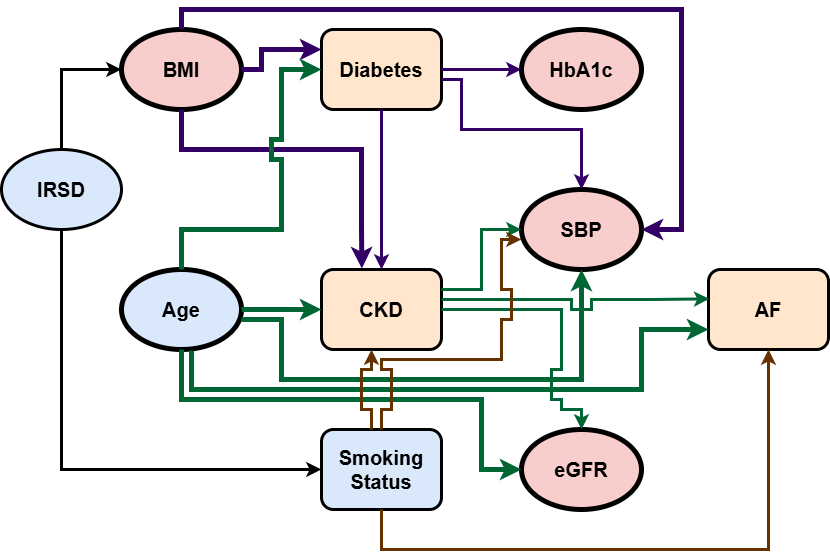}
    \caption{Directed acyclic graph representing the causal structure embedded in PRIME-CVD. Circular and rectangular nodes denote numeric and categorical variables, respectively; blue, orange, and red represent demographic/lifestyle determinants, chronic diseases, and anthropometric/physiological measurements.}
    \label{fig:DAG}
\end{figure}

All parameters are encoded within a directed acyclic graph that summarises the causal structure of the data-generating process (Figure \ref{fig:DAG}). Each parent-child relationship, together with its empirical justification, is listed in Table \ref{tab:DAG_edges}. Variable generation proceeds conditionally along this graph: nodes are sampled using logistic, linear, or mixture-based formulations calibrated to reproduce the epidemiologic quantities derived from these sources, ensuring internal coherence across the synthetic cohort. Full methodological details, including mathematical specifications and executable code for each component of the generative model, are provided in Appendix \textcolor{blue}{\textbf{A}}.

\textbf{Cohort Generation, Scale, and Data Flow.}\\
PRIME-CVD assembles a cohort of 50,000 adults aged 18-90\,years, reflecting the typical age range of primary-prevention cardiovascular studies and remaining consistent with the population distributions used to parameterise the model~\cite{rao2026transformer}. The cohort design deliberately mirrors contemporary EMR-based primary-prevention populations, enabling students to replicate key elements of recent 5-year absolute CVD risk-modelling studies. Importantly, Cox proportional hazards models~\cite{cox1972regression} fitted to the simulated cohort yield hazard ratio estimates that closely resemble those reported in contemporary Australian CVD studies~\cite{nicholas2025estimating}.

Since the data are fully simulated, no additional exclusion criteria are required; chronic conditions and biomarker values arise entirely from the probabilistic structure encoded in the DAG. Each individual is assigned a 5-year follow-up period, with continuous time-to-event quantities generated relative to a common baseline and subsequently mapped to calendar time to provide realistic event timing.

Construction of the clean, analysis-ready cohort (Data Asset 1) follows the DAG structure. First, exogenous variables (IRSD quintile, age) are sampled according to their target population distributions. Second, behavioural and anthropometric variables (smoking status, BMI) are generated conditional on IRSD, propagating socioeconomic gradients into health behaviours and adiposity. Third, chronic conditions (diabetes, CKD, AF) are assembled using logistic models whose coefficients derive from published odds ratios. Fourth, continuous biomarkers (HbA1c, SBP, eGFR) are drawn from linear or mixture-based models incorporating demographic, behavioural, metabolic, and renal influences. Finally, cardiovascular event times are simulated using a proportional-hazards model with a baseline hazard calibrated to yield an approximately 4\% five-year CVD incidence; individuals without events are administratively censored at the end of follow-up.

%%%%%%%%%%%%%%%%%%%%%%%%%%%%%%%%%%%%%%%%%%%%%%%%%%%%%%%%%%%%
%%%%%%%%%%%%%%%%%%%%%%%%%%%%%%%%%%%%%%%%%%%%%%%%%%%%%%%%%%%%
%%%%%%%%%%%%%%%%%%%%%%%%%%%%%%%%%%%%%%%%%%%%%%%%%%%%%%%%%%%%
%\newpage
\subsection{Data Asset 2: Divide, Distort, \& Disassemble}
\begin{figure}[h!]
    \centering
    \includegraphics[width=\linewidth]{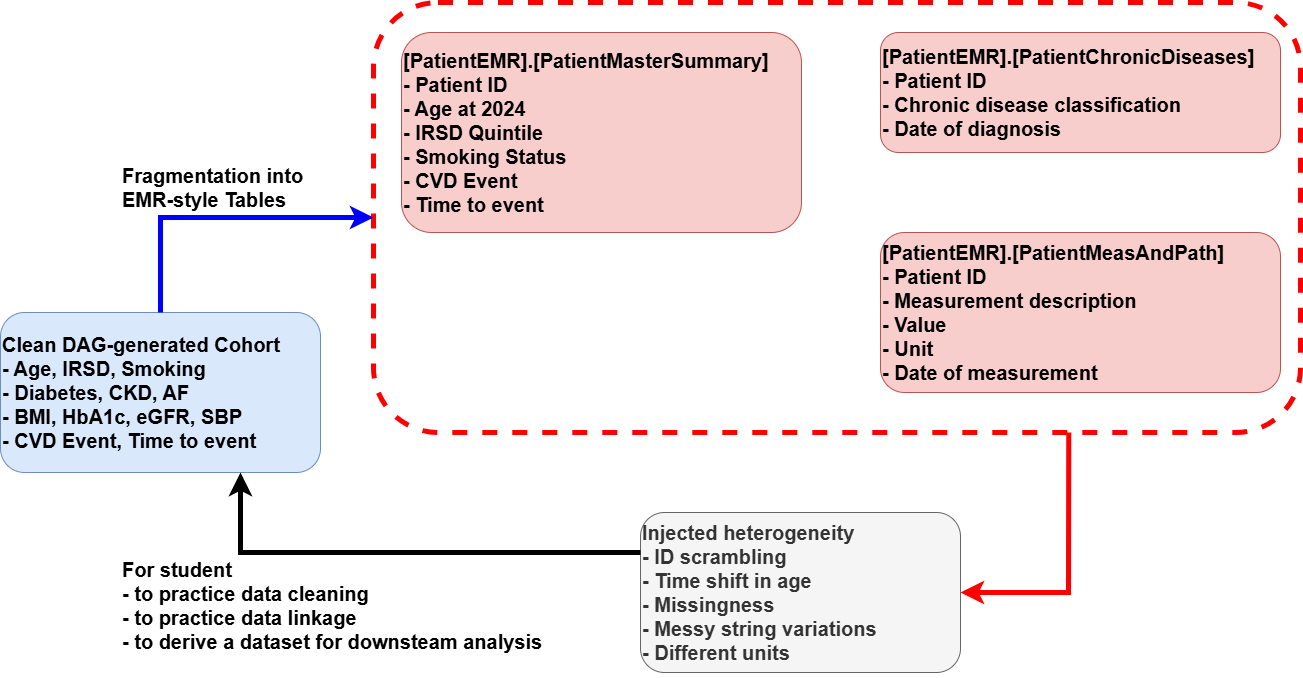}
    \caption{Constructing the PRIME-CVD EMR-style Data Asset 2. The clean DAG-generated cohort is split into three relational tables and augmented with realistic EMR artefacts including, missingness, ID scrambling, heterogeneous terminology, and mixed units.}
    \label{fig:Pipeline}
\end{figure}

The second stage transforms the clean DAG-generated cohort into PRIME-CVD Data Asset~2, a relational EMR-style dataset denoted \texttt{[PatientEMR]}. As summarised in Figure~\ref{fig:Pipeline}, all variables from Data Asset~1 are retained but redistributed across three linked tables and augmented with artefacts that mimic routinely collected general-practice and hospital records.

PRIME-CVD comprises three relational tables:\\ 
\hspace*{5mm}
\begin{minipage}{0.9\textwidth}
\texttt{[PatientEMR].[PatientMasterSummary]} contains one row per patient with a synthetic identifier, demographics, IRSD quintile, smoking status (with injected missingness), and coarsened CVD outcome timing; 
\end{minipage}\\
\hspace*{5mm}
\begin{minipage}{0.9\textwidth}
 \texttt{[PatientEMR].[PatientChronicDiseases]} expands baseline disease flags (diabetes, CKD, AF) into one-to-many diagnosis records using heterogeneous free-text labels and non-informative diagnosis months; and
\end{minipage}\\
\hspace*{5mm}
\begin{minipage}{0.9\textwidth}
 \texttt{[PatientEMR].[PatientMeasAndPath]} stores HbA1c, SBP, BMI, and eGFR as long-form measurement records with varied string descriptions and mixed HbA1c units. 
\end{minipage}\\
This decomposition mirrors common EMR architectures in which identifiers, problems, and measurements reside in distinct systems and must be re-linked for analysis. The transformation pipeline are implemented deterministically from Data Asset~1, ensuring that every artefact is reproducible. Full technical details of the relational construction are provided in Appendix~\textcolor{blue}{\textbf{B}}, where we provide code fragments supporting cohort reconstruction.

To emulate the challenges of working with real EMR data, we inject controlled “messiness’’ into all three tables: non-sequential but reproducible patient IDs; patterned missingness in smoking status; lexical heterogeneity in diagnosis and measurement labels; unit inconsistency for a subset of HbA1c results; and temporally diffused diagnosis and measurement dates that are decoupled from the baseline risk assessment window. Full technical details of the sampling rules are provided in Appendix~\textcolor{blue}{\textbf{C}}, where the distributions governing structural and lexical distortions are documented quantitatively.

%%%%%%%%%%%%%%%%%%%%%%%%%%%%%%%%%%%%%%%%%%%%%%%%%%%%%%%%%%%%
%%%%%%%%%%%%%%%%%%%%%%%%%%%%%%%%%%%%%%%%%%%%%%%%%%%%%%%%%%%%
%%%%%%%%%%%%%%%%%%%%%%%%%%%%%%%%%%%%%%%%%%%%%%%%%%%%%%%%%%%%
%\newpage
%------------------------------------------------------------------
\subsection{Deidentification, Ethics, and Governance}

Real EMR datasets pose several re-identification risks: direct identifiers (names, record numbers, full birth dates), quasi-identifiers~\cite{el2020evaluating} (precise timestamps, rare diagnoses, unique demographic combinations), and free-text notes that may contain protected health information. Synthetic datasets generated by machine learning models trained on real EMR data may still permit membership inference through generative-model inversion~\cite{kuo2022health,zhang2020secret}. These risks normally necessitate extensive de-identification pipelines and restrictive data-governance frameworks~\cite{UNSW_ERICA_2021}. Since PRIME-CVD contains no personal data and cannot be mapped to real individuals, it falls outside human-subjects research requirements, and no ethics approval was needed.

%------------------------------------------------------------------

\subsection{Code Availability}~\label{Sec:CodeAvail}

All code required to generate PRIME-CVD and reproduce the results in this manuscript will be made publicly available upon acceptance. This includes the DAG specification and parametrisation, cohort sampling functions, baseline-hazard calibration, and deterministic scripts for constructing the relational EMR-style dataset. The release will also include end-to-end notebooks to reproduce every table and figure.

Prior to release, Appendices~\textcolor{blue}{\textbf{A--C}} document the core generative mechanisms through code excerpts and mathematical details, while Appendix~\textcolor{blue}{\textbf{D}} provides reproducible notebooks demonstrating figure and table regeneration using the distributed data assets.

%%%%%%%%%%%%%%%%%%%%%%%%%%%%%%%%%%%%%%%%%%%%%%%%%%%%%%%%%%%%
%%%%%%%%%%%%%%%%%%%%%%%%%%%%%%%%%%%%%%%%%%%%%%%%%%%%%%%%%%%%
%%%%%%%%%%%%%%%%%%%%%%%%%%%%%%%%%%%%%%%%%%%%%%%%%%%%%%%%%%%%
\newpage
\section{Data Records}
The analysis ready cohort of Data Asset 1 contains 50,000 simulated patient data. A five-row preview of the asset is shown in Table~\ref{tab:da1_head} to illustrate variable types, typical ranges, and the joint structure students will encounter when fitting risk models. Data Asset 1 is provided as a single comma separated value (CSV) file (6.16 MB) containing one row per individual with the canonical variables used throughout the manuscript (IRSD quintile, age, smoking, BMI, diabetes, CKD, HbA1c, eGFR, SBP, AF, binary CVD event and continuous follow-up time). Users can load this file directly for model-building, exploratory visualisation (see Figure~\ref{fig:prime_baseline_dist}) or to reproduce the cohort summaries reported in Table~\ref{Table:Table01}.

The relational Data Asset 2 decomposes the same population into 3 EMR-style CSVs. The unified data dictionary in Table~\ref{Tab:Dic} describes each table and column. Three preview tables show the first five rows of each CSV: \texttt{PatientMasterSummary} (Table~\ref{tab:da2_master_head}; 1.81 MB), \texttt{PatientChronicDiseases} (Table~\ref{tab:da2_chronic_head}; 139 kB), and \texttt{PatientMeasAndPath} (Table~\ref{tab:da2_meas_head}; 10.40 MB). In practice, \texttt{PatientMasterSummary} provides the patient index, age at 2024, baseline smoking and IRSD, and coarsened CVD outcome; \texttt{PatientChronicDiseases} is a presence-only list with heterogeneous labels and diagnosis months; and \texttt{PatientMeasAndPath} is a long-form measurement table containing variable descriptions, mixed units, and dates. Every table uses the deterministic synthetic \texttt{Patient\_ID} for linkage; see Table~\ref{Tab:Dic} for column semantics and expected value formats.

All PRIME-CVD assets are released as versioned CSV files on FigShare
(\cite{kuo2026primecvd_da1, kuo2026primecvd_da2}).
The format ensures platform-agnostic ingestion and compatibility with
common analytical workflows in Python~\cite{python_software_foundation_2016}
and R~\cite{r2021r}. Reproducible quick-start notebooks are provided for both
environments\footnote{Refer to a quick start in Python via \url{https://github.com/NicKuo-ResearchStuff/PRIME_CVD/blob/main/2026_02_25_PrimeCvd_QuickStart.ipynb}.}\footnote{Refer to a quick start in R via \url{https://github.com/NicKuo-ResearchStuff/PRIME_CVD/blob/main/2026_02_25_PrimeCvd_QuickStart(R_Version).ipynb}.}, and the files can be executed without modification in
Google Colab and Jupyter Notebook settings.

%%%%%%%%%%%%%%%%%%%%%%%%%%%%%%%%%%%%%%%%%%%%%%%%%%%%%%%%%%%%
%%%%%%%%%%%%%%%%%%%%%%%%%%%%%%%%%%%%%%%%%%%%%%%%%%%%%%%%%%%%
%%%%%%%%%%%%%%%%%%%%%%%%%%%%%%%%%%%%%%%%%%%%%%%%%%%%%%%%%%%%
%\newpage
\section{Technical Validation}

Since PRIME-CVD is designed for medical education, this section adopts a pedagogical validation framework. We present three representative assignment-style exercises that reflect common learning objectives in health informatics training: EMR-style cohort reconstruction, socioeconomic stratified analysis, and multivariable time-to-event modelling, illustrating complementary pathways for developing practical, policy-relevant analytic competence.

\begin{figure}[h!]
    \centering
    \includegraphics[width=\linewidth]{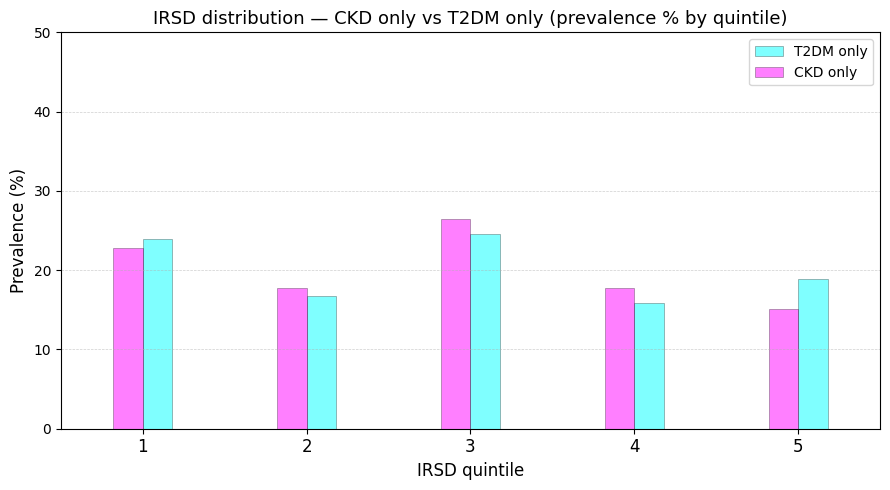}
    \caption{Socioeconomic distributions for mutually exclusive CKD and T2DM cohorts reconstructed from the relational Data Asset 2. Bar plots summarise the prevalence of IRSD quintiles within each cohort following linkage and harmonisation of diagnosis records, enabling comparison of socioeconomic profiles across conditions.}\label{fig:da2_cohort_irsd_comparison}
\end{figure}

\textbf{Exploratory cohort construction and socioeconomic visualisation.}\\
To validate the pedagogical utility of the relational EMR-style Data Asset 2, we present an exploratory cohort reconstruction exercise comparing patients with CKD and diabetes (T2DM). This task requires linking multiple tables, harmonising heterogeneous diagnosis labels, and reconstructing mutually exclusive CKD-only and T2DM-only cohorts from the EMR-style records. Students then summarise socioeconomic characteristics by visualising the distribution of IRSD quintiles within each cohort. A representative comparison is shown in Figure~\ref{fig:da2_cohort_irsd_comparison}. Full implementation details, including a reference solution and code used to derive this figure, are provided in Appendices~\textcolor{blue}{\textbf{D.1}} and \textcolor{blue}{\textbf{D.2}}.

\textbf{IRSD-stratified distributional checks.}\\
To support understanding of socioeconomic stratification, we present IRSD-stratified distributions of key demographic, behavioural, clinical, and outcome variables in Data Asset 1. These summaries demonstrate coherent gradients in risk factors and cardiovascular outcomes across deprivation strata. For students, this exercise illustrates how stratification reveals structure obscured in pooled analyses, informs calibration~\cite{van2019calibration}, and motivates the development of risk-prediction models that are not only accurate but also fair and equitable across the socioeconomic spectrum. Expected student-derived patterns are shown in Table~\ref{tab:prime_irsd} and Figure~\ref{fig:ZFig04_Table2IrsdStratification}, with further details provided in Appendices~\textcolor{blue}{\textbf{D.3}} and \textcolor{blue}{\textbf{D.4}}.

\begin{figure}[h!]
    \centering
    \includegraphics[width=\linewidth]{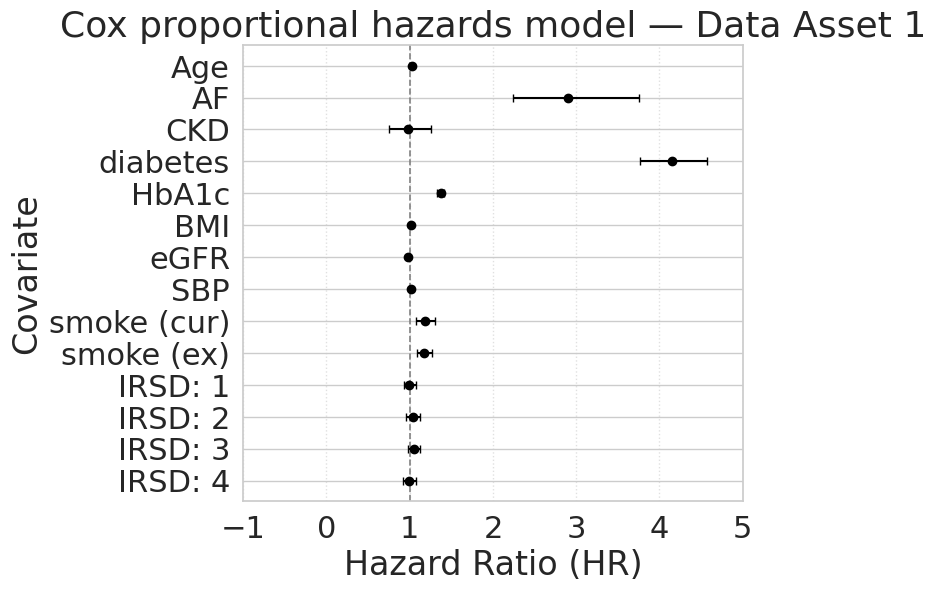}
    \caption{Hazard ratios derived from a Cox model fitted to Data Asset~1.}
    \label{fig:ZFig22}
\end{figure}

\textbf{Multivariable hazard modelling and policy-relevant interpretation.}\\
To consolidate understanding of time-to-event modelling using clean cohort data, students are expected to fit a multivariable Cox proportional hazards model~\cite{cox1972regression} to Data Asset~1 and derive adjusted hazard ratios for 5-year cardiovascular risk. The resulting table of hazard ratios (Table~\ref{tab:cox_hr_dataasset1}; baseline: non-smoker, IRSD quintile~5, Age at 30, no chronic disease) and the corresponding forest plot (Figure~\ref{fig:ZFig22}) illustrate the relative contribution of established risk factors. This exercise reinforces interpretation of adjusted effects and demonstrates how survival models can inform equitable, evidence-based cardiovascular risk policy. See further details, including a reference solution and code, provided in Appendices~\textcolor{blue}{\textbf{D.5}} and \textcolor{blue}{\textbf{D.6}}.

\textbf{Additional notes.}\\
More validation analyses, including complete correlation structure, age-stratified distributions, and a worked reconstruction pipeline for the relational asset, are provided in Appendix~\textcolor{blue}{\textbf{E}} to further document the intended epidemiologic structure of PRIME-CVD.

%%%%%%%%%%%%%%%%%%%%%%%%%%%%%%%%%%%%%%%%%%%%%%%%%%%%%%%%%%%%
%%%%%%%%%%%%%%%%%%%%%%%%%%%%%%%%%%%%%%%%%%%%%%%%%%%%%%%%%%%%
%%%%%%%%%%%%%%%%%%%%%%%%%%%%%%%%%%%%%%%%%%%%%%%%%%%%%%%%%%%%
%\newpage
\section{Usage Notes}\label{Sec:UsageNotes}
\textbf{Discussion.}\\
PRIME-CVD addresses a persistent tension in medical education and methodological training: the trade-off between privacy protection and analytic realism in EMR data. In real-world releases, rare but clinically important combinations of characteristics -- such as young adults with type 2 diabetes mellitus and aggressive kidney disease in specific ancestry groups~\cite{spanakis2013race} -- often form very small, high-risk strata. To mitigate re-identification risk, these strata are routinely pooled or suppressed, limiting opportunities to teach health inequities, subgroup risk, and fairness-aware modelling.

By generating all individuals, events, and covariates de novo from publicly available aggregate statistics and published epidemiologic effect estimates, PRIME-CVD retains realistic subgroup imbalance and risk gradients without exposing direct or quasi-identifiers. Because no synthetic record corresponds to a real individual, learners can interrogate clinically meaningful heterogeneity while maintaining strict privacy guarantees.

\textbf{Curricular integration and educational alignment.}\\
The two data assets serve complementary pedagogical functions. Data Asset~1 provides a structured cohort for regression, survival modelling, and calibration assessment of predicted 5-year cardiovascular risk, including stratified analyses across socioeconomic strata. Data Asset~2 reintroduces EMR-style structural and lexical heterogeneity, requiring table linkage, variable harmonisation, unit standardisation, and cohort reconstruction prior to analysis.

This design aligns with the CBDRH Health Data Science curriculum. In Data Management \& Curation (HDAT9400), students derive an analysis-ready dataset from relational EMR tables while documenting data dictionaries and quality-control procedures. In Statistical Modelling (HDAT9600/9700), the cohort supports generalised linear models and Cox proportional hazards modelling with diagnostic and calibration assessment. In Machine Learning and Visualisation (HDAT9500/9510, HDAT9800), the dataset enables dimensionality reduction (\textit{e.g.,} t-SNE~\cite{van2008visualizing}), subgroup performance evaluation, and comparative assessment of predictive models under known ground-truth structure.

PRIME-CVD is not a substitute for real clinical data, nor are models trained on it clinically deployable. Rather, it provides a reproducible environment in which methodological, computational, and translational components of health data science can be developed and critically evaluated prior to application in governed settings.

\textbf{Availability and community engagement.}\\
The official PRIME-CVD repository is available at\\ 
\url{https://github.com/NicKuo-ResearchStuff/PRIME_CVD},\\ 
where documentation, instructional blogs, and fully reproducible notebooks are maintained.

The repository illustrates direct programmatic access to the datasets. For example, Data Asset~1 can be loaded in Python as follows:

\begin{lstlisting}[language=Python, style=mystyle, backgroundcolor=\color{backcolour4Input}]
import pandas as pd

url = "https://ndownloader.figshare.com/files/62102364"
df = pd.read_csv(url)
df = df.drop(columns=["Unnamed: 0"], errors="ignore")
\end{lstlisting}

Accompanying notebooks demonstrate cohort reconstruction from the EMR-style relational tables, variable harmonisation and unit standardisation, and end-to-end modelling workflows. 
Lecturers and educators are free to adopt, adapt, or extend the provided materials when developing course content, classroom examples, or assessment tasks, including in disciplines beyond health data science. 
By combining openly accessible data with executable teaching resources, PRIME-CVD supports flexible and reproducible educational deployment without administrative access barriers.

%%%%%%%%%%%%%%%%%%%%%%%%%%%%%%%%%%%%%%%%%%%%%%%%%%%%%%%%%%%%
%%%%%%%%%%%%%%%%%%%%%%%%%%%%%%%%%%%%%%%%%%%%%%%%%%%%%%%%%%%%
%%%%%%%%%%%%%%%%%%%%%%%%%%%%%%%%%%%%%%%%%%%%%%%%%%%%%%%%%%%%
\newpage
\section*{Figures \& Tables}

%###===>>>%###===>>>%###===>>>
%###===>>>%###===>>>%###===>>>
%###===>>>%###===>>>%###===>>>
%\newpage
\begin{table}[h!]
\centering
\renewcommand{\arraystretch}{1.3}
\begin{tabular}{p{4.5cm} p{8cm}}
\hline
\multicolumn{2}{l}{\textbf{Synthetic Cohort Characteristics (N = 50{,}000)}} \\
\hline
\textbf{Age (years)} 
& Mean (SD): 49.71 (12.37) \\
& Median [IQR]: 49.63 [41.33, 58.09] \\
& Range: 18.0–90.0 \\

\hline
\textbf{IRSD Quintile Distribution} 
& Q1: 21.28\% (Most disadvantaged)\\
& Q2: 16.11\% \\
& Q3: 23.88\% \\
& Q4: 16.99\% \\
& Q5: 21.74\% (Least disadvantaged)\\

\hline
\textbf{Smoking Status} 
& Non-smoker: 73.14\% \\
& Ex-smoker: 16.72\% \\
& Current smoker: 10.13\% \\

\hline
\textbf{Chronic Disease Prevalence}
& Diabetes: 7.43\% \\
& Chronic Kidney Disease (CKD): 0.680\% \\
& Atrial Fibrillation (AF): 0.720\% \\

\hline
\textbf{Body Mass Index}
& Mean (SD): 28.33 (5.03) \\
\textbf{(BMI, kg/m$^2$)}& Median [IQR]: 28.33 [24.92, 31.73] \\
& Range: 15.0–52.76 \\

\hline
\textbf{Systolic Blood Pressure} 
& Mean (SD): 123.31 (16.10) \\
\textbf{(SBP, mmHg))} & Median [IQR]: 123.14 [112.39, 134.10] \\
& Range: 55.85–187.79 \\

\hline
\textbf{Estimated Glomerular}
& Mean (SD): 82.77 (6.09) \\
\textbf{Filtration Rate} & Median [IQR]: 82.94 [79.22, 86.66] \\
\textbf{(eGFR, mL/min/1.73m$^2$)} & Range: 37.00–104.65 \\

\hline
\textbf{Haemoglobin A1c} 
& Mean (SD): 4.79 (0.93) \\
\textbf{(HbA1c, \%)} & Median [IQR]: 4.66 [4.24, 5.12] \\
& Range: 2.23–12.71 \\

\hline
\hline
\textbf{Cardiovascular Outcomes}
& Overall CVD event rate: 4.02\% \\
\textbf{Follow-up Time (years)}
& Mean: 4.80 \\

\hline
\end{tabular}
\caption{Baseline characteristics of the PRIME-CVD cohort used for cardiovascular risk simulation.}
\label{Table:Table01}
\end{table}

%###===>>>%###===>>>%###===>>>
%###===>>>%###===>>>%###===>>>
%###===>>>%###===>>>%###===>>>
\newpage

%%%%%%%%%%%%%%%%%%%%%%%%%%%%%%%%%%%%%%%%%%%%%%%%%%%%%%%%%%%%
%%%%%%%%%%%%%%%%%%%%%%%%%%%%%%%%%%%%%%%%%%%%%%%%%%%%%%%%%%%%
%%%%%%%%%%%%%%%%%%%%%%%%%%%%%%%%%%%%%%%%%%%%%%%%%%%%%%%%%%%%
%\newpage
% FIGURE 1 — DAG

%###===>>>%###===>>>%###===>>>
%###===>>>%###===>>>%###===>>>
%###===>>>%###===>>>%###===>>>
\newpage
\begin{table}[h!]
\footnotesize
\centering
\caption{Unified data dictionary for the PRIME-CVD relational dataset \texttt{[PatientEMR]}.\label{Tab:Dic}}
\begin{tabular}{p{3.5cm} p{2.5cm} p{6.5cm}}
\hline
\textbf{Table} & \textbf{Column Name} & \textbf{Notes} \\
\hline
\hline

\texttt{PatientMasterSummary} 
& Patient\_ID 
& Patient identifier:\newline
Unique, non-sequential integer key derived from the cohort row index of Data Asset 1. 
\\ \\

\texttt{PatientMasterSummary} 
& Age\_At\_2024 
& Approximate patient age in 2024:\newline 
Continuous, two-decimal precision, of baseline\newline simulated age exacted at 2024. 
\\ \\

\texttt{PatientMasterSummary} 
& IRSD\_Quintile 
& Australian socioeconomic disadvantage quintile:\newline
Integer in \{1,...,5\}, where 1 = most disadvantaged and 5 = least disadvantaged. 
\\ \\

\texttt{PatientMasterSummary} 
& SMOKING\_STATUS 
& Baseline smoking category:\newline 
Categorical: \texttt{"non"}, \texttt{"current"}, \texttt{"ex"}, \texttt{"N/A"}.
\\ \\

\texttt{PatientMasterSummary} 
& CVD\_Event 
& Occurrence of composite CVD outcome:\newline
Binary: 1 = event during follow-up; 0 = censored. 
\\ \\

\texttt{PatientMasterSummary} 
& CVD\_Time 
& Calendar year–month of event or censoring:\newline 
\texttt{YYYY-MM} format; events mapped from simulated time-to-event since 2017-01-01;\newline
non-events censored at \texttt{"2022-12"}. 
\\ \\

\hline
\hline
\texttt{PatientChronicDiseases} 
& Patient\_ID 
& Patient identifier:\newline 
Links to \texttt{PatientMasterSummary}; multiple rows per patient possible. 
\\ \\

\texttt{PatientChronicDiseases} 
& Classification 
& Free-text chronic disease label:\newline
Heterogeneous terminology describing diabetes, CKD, or AF (\textit{e.g.}, ICD codes, abbreviations,\newline verbose descriptions). 
\\ \\

\texttt{PatientChronicDiseases} 
& Date 
& Diagnosis year–month:\newline 
\texttt{YYYY-MM} format. 
\\ \\

\hline
\hline

\texttt{PatientMeasAndPath} 
& Patient\_ID 
& Patient identifier:\newline 
Links to \texttt{PatientMasterSummary}. 
\\ \\

\texttt{PatientMeasAndPath} 
& Description  
& Canonical biomarker name:\newline
One of \texttt{"HbA1c"}, \texttt{"eGFR"}, \texttt{"SBP"} but with heterogeneous string variants (abbreviations, case differences, LOINC-like codes). 
\\ \\

\texttt{PatientMeasAndPath} 
& Value 
& Numeric measurement value:\newline 
Interpretation depends on the measurement type and associated unit. 
\\ \\

\texttt{PatientMeasAndPath} 
& Unit 
& Measurement unit:\newline 
\texttt{"mmHg"} for SBP, \texttt{"mL/min/1.73m²"} for eGFR,\newline 
\texttt{"\%"} or \texttt{"mmol/mol"} for HbA1c. 
\\ \\

\texttt{PatientMeasAndPath} 
& Date 
& Measurement year–month:\newline 
\texttt{YYYY-MM} format. 
\\ \\

\hline
\end{tabular}
\end{table}

%%%%%%%%%%%%%%%%%%%%%%%%%%%%%%%%%%%%%%%%%%%%%%%%%%%%%%%%%%%%
%%%%%%%%%%%%%%%%%%%%%%%%%%%%%%%%%%%%%%%%%%%%%%%%%%%%%%%%%%%%
%%%%%%%%%%%%%%%%%%%%%%%%%%%%%%%%%%%%%%%%%%%%%%%%%%%%%%%%%%%%
\newpage
% TABLE 1 — Simplified DAG Edges
\begin{table}[h!]
\centering
\caption{Simplified Parent--Child Relationships in the PRIME-CVD DAG}\label{tab:DAG_edges}
\begin{tabular}{p{3cm} p{7.5cm} p{2cm}}
\hline
\textbf{Parent → Child} & \textbf{Notes} & \textbf{References} \\

\hline
\hline
IRSD $\rightarrow$ BMI & Higher deprivation slightly increases BMI. &
\cite{From_IRSD_to_BMI_AIHW_OverweightObesityReport}$^{\dagger}$\cite{From_IRSD_to_BMI_AIHW_InequalitiesOverweight}$^{\dagger}$ \\\\

IRSD $\rightarrow$ Smoking & More disadvantaged areas have higher smoking rates. &
\cite{From_IRSD_to_Smoking_AIHW_NDSHS}$^{\dagger}$\cite{From_IRSD_to_Smoking_TobaccoInAustralia}$^{\star}$ \\\\

\hline
\hline
BMI $\rightarrow$ Diabetes & Higher BMI increases diabetes risk. &
\cite{From_BMI_to_T2DM_and_SBP_NHMRC_ObesityReview}$^{\dagger}$\cite{From_BMI_to_T2DM_and_SBP_ObesityEvidenceHub}$^{\star}$ \\\\

BMI $\rightarrow$ CKD & Higher BMI mildly increases CKD probability. &
\cite{From_Age_to_CKD_AIHW_Prevalence}$^{\dagger}$\cite{From_BMI_to_CKD_FirstAuthorEtAl2021_OSP4_629}$^{\diamond}$ \\\\

BMI $\rightarrow$ SBP & Higher BMI increases SBP. &
\cite{From_BMI_to_T2DM_and_SBP_NHMRC_ObesityReview}$^{\dagger}$\cite{From_BMI_to_T2DM_and_SBP_ObesityEvidenceHub}$^{\star}$ \\\\

\hline
\hline
Age $\rightarrow$ Diabetes & Diabetes becomes more common with age. &
\cite{From_Age_to_Diabetes_AIHW_AllDiabetes}$^{\dagger}$\cite{From_Age_to_Diabetes_ABS_LatestRelease}$^{\dagger}$ \\\\

Age $\rightarrow$ CKD & CKD prevalence increases with age. &
\cite{From_Age_to_CKD_AIHW_Prevalence}$^{\dagger}$\cite{From_Age_to_CKD_RACGP_CKDInTheElderly}$^{\star}$ \\\\

Age $\rightarrow$ SBP & Blood pressure rises with age. &
\cite{From_Age_to_SBP_AIHW_Summary}$^{\dagger}$\cite{From_Age_to_SBP_ABS_HypertensionLatest}$^{\dagger}$ \\\\

Age $\rightarrow$ eGFR & Kidney function declines with age. &
\cite{From_Age_to_CKD_AIHW_Prevalence}$^{\dagger}$ \\\\

Age $\rightarrow$ AF & AF becomes more common with age. &
\cite{From_Age_to_AF_AIHW_Facts}$^{\dagger}$\cite{From_Age_to_AF_FirstAuthorEtAl2021_PMC8330568}$^{\diamond}$ \\\\

\hline
\hline
Diabetes $\rightarrow$ CKD & Diabetes substantially increases CKD risk. &
\cite{From_Diabetes_and_CKD_and_AF_FirstAuthorEtAl2023_JAHA028496}$^{\diamond}$ \\\\

Diabetes $\rightarrow$ HbA1c & Diabetes increases glycaemia levels. &
\cite{From_Diabetes_to_HbA1c_RACGP_MoreThanJustANumber}$^{\star}$\cite{From_Diabetes_to_HbA1c_ADA_A1c}$^{\circ}$ \\\\

Diabetes $\rightarrow$ SBP & Diabetes elevates SBP. &
\cite{From_Diabetes_to_SBP_FirstAuthorEtAl_PMC12278073}$^{\diamond}$\cite{From_Diabetes_to_SBP_VerywellHealth_DiabetesLowBP}$^{\circ}$ \\\\

\hline
\hline
CKD $\rightarrow$ SBP & CKD strongly elevates SBP. &
\cite{From_CKD_to_SBP_FirstAuthorEtAl2023_NDT}$^{\diamond}$ \\\\

CKD $\rightarrow$ eGFR & CKD markedly lowers eGFR. &
\cite{From_CKD_to_eGFR_FirstAuthorEtAl2025_EurHeartJ}$^{\diamond}$\cite{From_CKD_to_eGFR_and_AF_FirstAuthorEtAl2022_PMC9445413}$^{\diamond}$ \\\\

CKD $\rightarrow$ AF & CKD strongly increases AF risk. &
\cite{From_CKD_to_eGFR_and_AF_FirstAuthorEtAl2022_PMC9445413}$^{\diamond}$ \\\\

\hline
\hline
Smoking $\rightarrow$ CKD & Smokers have higher CKD risk. &
\cite{From_Smoking_to_CKD_FirstAuthorEtAl2017_PMID28339863}$^{\diamond}$\cite{From_Smoking_to_CKD_FirstAuthorEtAl2010_BMC}$^{\diamond}$ \\\\

Smoking $\rightarrow$ SBP & Smokers have slightly higher SBP. &
\cite{From_Smoking_to_SBP_AIHW_BiomedicalRiskFactors}$^{\dagger}$ \\\\

Smoking $\rightarrow$ AF & Smokers have increased AF risk. &
\cite{From_Smoking_to_AF_FirstAuthorEtAl2018_PMID29996680}$^{\diamond}$\cite{From_Smoking_to_AF_FirstAuthorEtAl2017_PMC5407172}$^{\diamond}$ \\\\

\hline
\multicolumn{3}{l}{\footnotesize
$^{\dagger}$ Australian governmental agencies}\\
\multicolumn{3}{l}{\footnotesize
$^{\star}$ Australian health/medical organisations or websites} \\
\multicolumn{3}{l}{\footnotesize
$^{\diamond}$ Openly available peer-reviewed research}\\
\multicolumn{3}{l}{\footnotesize
$^{\circ}$ Openly available international resources} \\
\hline
\end{tabular}
\end{table}

%%%%%%%%%%%%%%%%%%%%%%%%%%%%%%%%%%%%%%%%%%%%%%%%%%%%%%%%%%%%
%%%%%%%%%%%%%%%%%%%%%%%%%%%%%%%%%%%%%%%%%%%%%%%%%%%%%%%%%%%%
%%%%%%%%%%%%%%%%%%%%%%%%%%%%%%%%%%%%%%%%%%%%%%%%%%%%%%%%%%%%
\newpage
% FIGURE 2 — DATA ASSET 2 PIPELINE

%%%%%%%%%%%%%%%%%%%%%%%%%%%%%%%%%%%%%%%%%%%%%%%%%%%%%%%%%%%%
%\newpage
\begin{table}[h!]
\centering
\scriptsize
\caption{First 5 rows from Data Asset~1.}
\label{tab:da1_head}
\begin{tabular}{llllllllllll}
\toprule
IRSD &
Age &
Smoking &
BMI &
Diabetes &
CKD &
HbA1c &
eGFR &
SBP &
AF &
CVD\_Event &
CVD\_Time \\
\midrule
4 & 50.40 & non & 28.17 & 0 & 0 & 4.32 & 83.08 & 118.67 & 0 & 1 & 2.96 \\
3 & 39.23 & ex  & 16.83 & 0 & 0 & 4.70 & 81.49 & 123.72 & 0 & 0 & 4.41 \\
5 & 55.49 & non & 23.52 & 0 & 0 & 3.67 & 86.78 & 126.65 & 0 & 0 & 4.49 \\
4 & 51.91 & non & 31.98 & 0 & 0 & 4.49 & 94.70 & 113.89 & 0 & 0 & 4.89 \\
1 & 47.09 & ex  & 25.35 & 0 & 0 & 4.32 & 86.34 & 125.91 & 0 & 0 & 4.49 \\
\bottomrule
\end{tabular}
\end{table}

\begin{table}[h!]
\centering
\scriptsize
\caption{First 5 rows from Data Asset~2's \texttt{PatientMasterSummary.csv}.}
\label{tab:da2_master_head}
\begin{tabular}{lllllll}
\toprule
Patient\_ID &
Age\_At\_2024 &
Smoking &
IRSD &
CVD\_Event &
CVD\_Time \\
\midrule
269 & 57.40 & non & 4 & 1 & 2019-12 \\
272 & 46.23 & ex  & 3 & 0 & 2022-12 \\
281 & 62.49 & non & 5 & 0 & 2022-12 \\
296 & 58.91 & non & 4 & 0 & 2022-12 \\
317 & 54.09 & ex  & 1 & 0 & 2022-12 \\
\bottomrule
\end{tabular}
\end{table}

\begin{table}[h!]
\centering
\footnotesize
\caption{First 5 rows from Data Asset~2's \texttt{PatientChronicDiseases.csv}.}
\label{tab:da2_chronic_head}
\begin{tabular}{lll}
\toprule
Patient\_ID &
Category &
Date \\
\midrule
1338163469 & Diabetes      & 2012-05 \\
150591944  & Diabetes      & 2014-12 \\
545940569  & ICD10: E11    & 2016-10 \\
7134075944 & Diabetes      & 2013-02 \\
5959722392 & Diabetes      & 2014-09 \\
\bottomrule
\end{tabular}
\end{table}

\begin{table}[h!]
\centering
\footnotesize
\caption{First 5 rows from Data Asset~2's \texttt{PatientMeasAndPath.csv}.}
\label{tab:da2_meas_head}
\begin{tabular}{llllll}
\toprule
Patient\_ID &
Measure &
Value &
Description &
Date &
Unit \\
\midrule
5386159421 & eGFR  & 77.24 & eGFR     & 2014-12 & mL/min/1.73m$^2$ \\
5790852944 & HbA1c & 4.50  & HBA1C    & 2013-02 & \% \\
3141314912 & HbA1c & 4.16  & Hb A1c   & 2016-09 & \% \\
5573606096 & eGFR  & 89.52 & eGFR     & 2013-10 & mL/min/1.73m$^2$ \\
3311104121 & SBP   & 131.55& SBP      & 2013-11 & mmHg \\
\bottomrule
\end{tabular}
\end{table}

%%%%%%%%%%%%%%%%%%%%%%%%%%%%%%%%%%%%%%%%%%%%%%%%%%%%%%%%%%%%
\newpage

%%%%%%%%%%%%%%%%%%%%%%%%%%%%%%%%%%%%%%%%%%%%%%%%%%%%%%%%%%%%
\newpage
\begin{table}[h!]
\centering
\scriptsize
\renewcommand{\arraystretch}{1.25}
\begin{tabular}{p{0.50cm} 
                p{0.60cm} 
                p{1.50cm} 
                p{0.70cm} 
                p{0.40cm} 
                p{0.40cm} 
                p{1.15cm} 
                p{1.15cm} 
                p{1.15cm} 
                p{1.15cm} 
                p{0.50cm} }
\hline
\textbf{IRSD} & 
\textbf{N} & 
\textbf{Age\newline median [IQR]} & 
\textbf{Diabetes\newline (\%)} & 
\textbf{CKD\newline (\%)} & 
\textbf{AF\newline (\%)} &
\textbf{BMI\newline mean (SD)} &
\textbf{SBP\newline mean (SD)} &
\textbf{eGFR\newline mean (SD)} &
\textbf{HbA1c\newline mean (SD)} &
\textbf{Event\newline (\%)} \\
\hline

1 & 
10{,}640 & 
49.53\newline[41.3–58.09] & 
8.44 & 0.77 & 0.81 & 
29.50\newline(4.95) & 
124.98\newline(16.12) & 
82.72\newline(6.22) & 
4.81\newline(0.97) & 4.28 \\\\

2 & 
8{,}055 & 
49.69\newline[41.44–58.26] & 
7.78 & 0.72 & 0.72 & 
28.99\newline(4.96) & 
124.23\newline(16.11) & 
82.75\newline(6.10) & 
4.80\newline(0.92) & 4.38 \\\\

3 & 
11{,}941 & 
49.54\newline[41.35–58.00] & 
7.53 & 0.80 & 0.69 & 
28.39\newline(4.95) & 
123.32\newline(16.21) & 
82.76\newline(6.14) & 
4.79\newline(0.93) & 4.34 \\\\

4 & 
8{,}495 & 
49.87\newline[41.59–58.23] & 
6.98 & 0.67 & 0.72 & 
27.74\newline(5.01) & 
122.57\newline(16.04) & 
82.72\newline(6.07) & 
4.77\newline(0.92) & 3.73 \\\\

5 & 
10{,}869 & 
49.62\newline[41.11–57.88] & 
6.44 & 0.44 & 0.67 & 
27.09\newline(4.92) & 
121.58\newline(15.80) & 
82.87\newline(5.93) & 
4.75\newline(0.91) & 3.37 \\\\

\hline
\end{tabular}
\caption{Characteristics of Data Asset 1 stratified by IRSD quintile.}
\label{tab:prime_irsd}
\end{table}

\begin{figure}[h!]
    \centering
    \includegraphics[width=\linewidth]{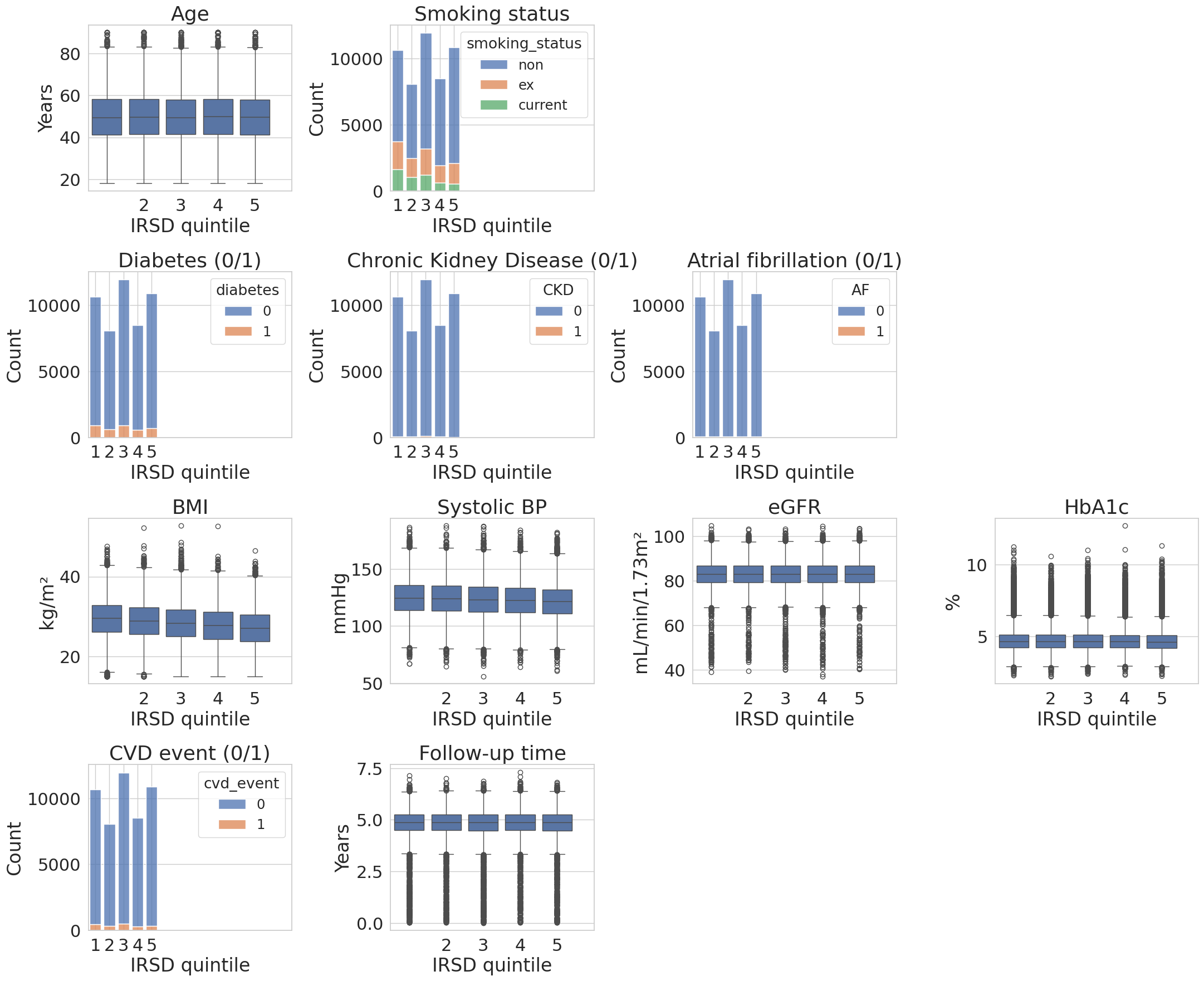}
    \caption{IRSD-stratified distributions of variables in Data Asset 1.}
    \label{fig:ZFig04_Table2IrsdStratification}
\end{figure}

%%%%%%%%%%%%%%%%%%%%%%%%%%%%%%%%%%%%%%%%%%%%%%%%%%%%%%%%%%%%
\newpage
\begin{table}[htbp]
\centering
\renewcommand{\arraystretch}{1.2}
\caption{Hazard ratios from a multivariable Cox proportional hazards model fitted to Data Asset~1.}
\label{tab:cox_hr_dataasset1}
\begin{tabular}{p{3.5cm} p{2cm} p{3cm} p{3cm}}
\toprule
\textbf{Covariate} & \textbf{HR} & \textbf{95\% CI} & \textbf{\textit{p}-value} \\
\midrule
Age (per year)          & 1.03 & [1.03, 1.03] & $<0.001$ \\
Atrial fibrillation     & 2.90 & [2.24, 3.75] & $<0.001$ \\
Chronic kidney disease  & 0.97 & [0.75, 1.26] & 0.84 \\
Diabetes mellitus       & 4.15 & [3.77, 4.57] & $<0.001$ \\
HbA1c (\%)              & 1.37 & [1.33, 1.41] & $<0.001$ \\
Body mass index         & 1.01 & [1.01, 1.02] & $<0.001$ \\
eGFR                    & 0.98 & [0.98, 0.99] & $<0.001$ \\
Systolic blood pressure & 1.01 & [1.01, 1.01] & $<0.001$ \\
Smoking (current)       & 1.18 & [1.08, 1.30] & $<0.001$ \\
Smoking (ex)            & 1.17 & [1.08, 1.26] & $<0.001$ \\
IRSD quintile 1         & 1.00 & [0.93, 1.07] & 0.92 \\
IRSD quintile 2         & 1.04 & [0.96, 1.13] & 0.36 \\
IRSD quintile 3         & 1.05 & [0.98, 1.12] & 0.19 \\
IRSD quintile 4         & 0.99 & [0.91, 1.07] & 0.82 \\
\bottomrule
\multicolumn{4}{p{11.5cm}}{Note: Hazard ratios are reported relative to the following reference categories:\newline \hspace*{9mm}non-smoker, IRSD quintile~5 (least disadvantaged), age 30~years, and\newline \hspace*{9mm}absence of chronic disease.}
\end{tabular}
\end{table}

%###===>>>%###===>>>%###===>>>
%###===>>>%###===>>>%###===>>>
%###===>>>%###===>>>%###===>>>
\newpage
\section*{Abbreviations}

\begin{table}[h!]
\centering
\begin{tabular}{p{2.5cm} p{10cm}}
\toprule
\textbf{Abbreviation} & \textbf{Definition} \\
\midrule
ABS & Australian Bureau of Statistics \\
AF & Atrial Fibrillation \\
AIHW & Australian Institute of Health and Welfare \\
BMI & Body Mass Index \\
CKD & Chronic Kidney Disease \\
CKM & Cardio–Kidney–Metabolic (domain including CVD, diabetes, and CKD) \\
CSV & Comma Separated Value \\
CVD & Cardiovascular Disease \\
DAG & Directed Acyclic Graph \\
DDPM & Denoising Diffusion Probabilistic Models \\
eGFR & Estimated Glomerular Filtration Rate \\
EMR & Electronic Medical Record \\
GAN & Generative Adversarial Network \\
HbA1c & Glycated Haemoglobin A1c \\
HR & Hazard Ratio \\
IRSD & Index of Relative Socioeconomic Disadvantage \\
NHS & National Health Service \\
NSW & New South Wales \\
OR & Odds Ratio \\
PRIME-CVD & Parametrically Rendered Informatics Medical Environment for\newline Cardiovascular Disease \\
SBP & Systolic Blood Pressure \\
T2DM & Type 2 Diabetes Mellitus \\
\bottomrule
\end{tabular}
\end{table}

%###===>>>%###===>>>%###===>>>
%###===>>>%###===>>>%###===>>>
%###===>>>%###===>>>%###===>>>
\newpage
\bibliographystyle{IEEEtran}
\bibliography{iclr2023_conference}

%###===>>>%###===>>>%###===>>>
%###===>>>%###===>>>%###===>>>
%###===>>>%###===>>>%###===>>>
\newpage
\appendix

{\LARGE \textbf{Appendix}}\\[10mm]

{\Large Supplementary Material for}\\[5mm]

{\Large \textbf{PRIME-CVD:}}\\[3mm]
{\Large A Parametrically Rendered Informatics Medical Environment}\\[2mm]
{\Large for Education in Cardiovascular Risk Modelling}

\vspace{15mm}

{\normalsize
Nicholas I-Hsien Kuo\\
Centre for Big Data Research in Health (CBDRH)\\
The University of New South Wales\\
Sydney, Australia
}

\vspace{5mm}

{\normalsize
Corresponding author: \texttt{n.kuo@unsw.edu.au}
}

\vfill

{\normalsize
This appendix constitutes the Supplementary Material accompanying the main manuscript.\\
It is intended to be read in conjunction with the primary paper.
}

%###===>>>%###===>>>%###===>>>
%###===>>>%###===>>>%###===>>>
%###===>>>%###===>>>%###===>>>
\newpage
\section{Detailed Assembly DAG Model and Parameterisation\newline(Data Asset 1 Implementation)}\label{App:A01}

This section details all covariates, biomarkers, and clinical conditions used to configure the causal DAG for PRIME-CVD Data Asset 1. 

%%%%%%%%%%%%%%%%%%%%%%%%%%%%%%%%%%%%%%%%%%%%%%%%%%%%%%%%%%%%
% FIGURE 1 — DAG
%%%%%%%%%%%%%%%%%%%%%%%%%%%%%%%%%%%%%%%%%%%%%%%%%%%%%%%%%%%%
\begin{figure}[h!]
    \centering
    \includegraphics[width=\linewidth]{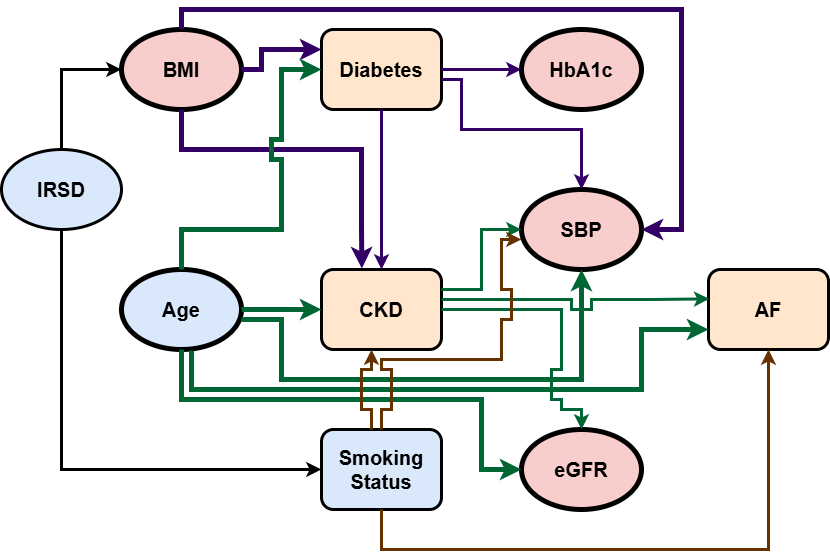}
    \caption{\newline Directed acyclic graph representing the causal structure embedded in PRIME-CVD:\newline 
    \hspace*{5mm} circular nodes: numeric vairables;\newline 
    \hspace*{5mm} rectangular nodes: binary/categorical variables;\newline 
    \hspace*{5mm} blue: demographic/lifestyle determinants;\newline 
    \hspace*{5mm} orange: chronic diseases;\newline
    \hspace*{5mm} red: anthropometric/physiological measurements.}
\end{figure}

\newpage
%###===>>>%###===>>>%###===>>>
%###===>>>%###===>>>%###===>>>
%###===>>>%###===>>>%###===>>>
\begin{table}[htbp]
\centering
\caption{PRIME-CVD DAG Edges -- Part 1}
\begin{tabular}{p{2cm} p{2cm} p{8.5cm}}
\hline
\textbf{From Node} & \textbf{To Node} & \textbf{Notes} \\ \hline

IRSD & SmokingStatus &
-- Each IRSD quintile uses a different probability vector over \texttt{["non", "current", "ex"]}.\newline
-- Higher deprivation (lower IRSD) has higher probability of being a current smoker.\newline
-- Implemented in \texttt{sample\_smoking\_given\_irsd()} via \texttt{rng.choice}. \\\\

IRSD & BMI &
-- BMI mean shifts linearly with $(Q_{\text{center}} - q)$, slope $\Delta_{\mathrm{BMI}}=0.6$.\newline
-- Central mean chosen so IRSD-weighted mean equals population BMI mean (28.29).\newline
-- Sampled from Normal$(\mu_q,\, \mathrm{SD}=4.98)$ with truncation to [15, 60]. \\\\

Age & Diabetes &
-- Diabetes log-odds uses $(\text{Age} - \text{AGE\_MEAN}) / 10$.\newline
-- Age coefficient $\log(1.8)$ encodes OR 1.8 per 10 years.\newline
-- Older age increases Bernoulli probability in \texttt{sample\_diabetes\_given\_age\_bmi()}. \\\\

BMI & Diabetes &
-- BMI enters as $(\text{BMI} - \text{BMI\_MEAN})$.\newline
-- Coefficient $\log(1.7)/5$ gives OR 1.7 per +5 BMI units.\newline
-- Higher BMI increases diabetes log-odds. \\\\

Age & CKD &
-- CKD log-odds uses $(\text{Age} - \text{AGE\_MEAN})$.\newline
-- Slope $\log(1.4)/10$ gives OR 1.4 per 10 years.\newline
-- Older age raises CKD probability. \\\\

Diabetes & CKD &
-- Diabetes adds a log-odds shift $\log(3.0)$.\newline
-- Implements OR approx 3 for CKD among diabetics.\newline
-- Applied as \texttt{BETA\_DM\_CKD * diabetes} in \texttt{sample\_ckd()}. \\\\

SmokingStatus & CKD &
-- Ex-smokers: $\log(1.2)$ increment; Current smokers: $\log(1.4)$.\newline
-- Implements OR 1.2 and OR 1.4 vs non-smokers.\newline
-- Added to linear predictor via status-specific masks. \\\\

BMI & CKD &
-- BMI enters as $(\text{BMI} - \text{BMI\_MEAN})$.\newline
-- Coefficient $\log(1.5)/5$ gives OR 1.5 per +5 BMI.\newline
-- Higher BMI increases CKD odds. \\\\

Diabetes & HbA1c &
-- HbA1c sampled from two-component mixture.\newline
-- Non-DM: Normal(4.6, 0.6). DM: Normal(7.1, 1.2).\newline
-- Diabetes determines which Gaussian is used. \\\\

Age & eGFR &
-- Age modifies expected eGFR via slope $-0.18$.\newline
-- Represents approx 1.8 mL/min/1.73m$^2$ decline per decade.\newline
-- Implemented as $(\text{Age} - \text{AGE\_MEAN})$ term. \\\\

\hline
\end{tabular}
\end{table}

\newpage
\begin{table}[htbp]
\centering
\caption{PRIME-CVD DAG Edges -- Part 2}
\begin{tabular}{p{2cm} p{2cm} p{8.5cm}}
\hline
\textbf{From Node} & \textbf{To Node} & \textbf{Notes} \\ \hline

CKD & eGFR &
-- CKD adds fixed shift $\Delta_{\mathrm{CKD}} = -31$.\newline
-- Strong downward deviation in filtration rate for CKD.\newline
-- Final eGFR $\sim$ Normal(mean, SD=5). \\\\

Age & SBP &
-- SBP mean includes age-centered term.\newline
-- Coefficient $0.38$ mmHg per year ($\approx 3.8$ per decade).\newline
-- Older age increases expected SBP. \\\\

SmokingStatus & SBP &
-- Ex-smokers: $+1.7$ mmHg; Current smokers: $+3.8$ mmHg.\newline
-- Added through \texttt{shift\_smoke} vector.\newline
-- Non-smokers are reference with shift=0. \\\\

BMI & SBP &
-- BMI effect: $1.0$ mmHg per BMI unit above mean.\newline
-- Linear contribution implemented via $(\text{BMI} - \text{BMI\_MEAN})$.\newline
-- Adiposity raises mean SBP. \\\\

Diabetes & SBP &
-- Diabetes adds fixed $+8$ mmHg to SBP mean.\newline
-- Implements DM-associated hypertension shift.\newline
-- Added as \texttt{DELTA\_SBP\_DM * diabetes}. \\\\

CKD & SBP &
-- CKD adds fixed $+12$ mmHg.\newline
-- Reflects strong renal–blood-pressure coupling.\newline
-- Implemented via \texttt{DELTA\_SBP\_CKD * ckd}. \\\\

Age & AF &
-- AF log-odds uses centered age.\newline
-- Coefficient $\log(1.7)/10$ gives OR 1.7 per decade.\newline
-- Older age increases AF probability. \\\\

SmokingStatus & AF &
-- Ex-smokers: $\log(1.1)$; Current: $\log(1.3)$.\newline
-- OR 1.1 and OR 1.3 vs non-smokers.\newline
-- Implemented by status-specific mask updates. \\\\

CKD & AF &
-- CKD adds log-odds shift $\log(3.0)$.\newline
-- approx 3-fold increase in AF odds for CKD individuals.\newline
-- Added via \texttt{BETA\_CKD\_AF * ckd}. \\\\

\hline
\end{tabular}
\end{table}

\newpage
\subsection{IRSD Quintile Sampling}\label{App:A12}
To model socioeconomic variation in baseline cardiovascular risk, we parameterise IRSD as a categorical latent variable taking values in quintiles $(1, \ldots, 5)$ each representing an ordered deprivation stratum. Empirical IRSD population frequencies are encoded directly via a fixed probability vector $(0.2136,0.1622,0.2393,0.1678,0.2171)$. For a cohort of size $n$, we generate IRSD assignments by sampling independently from this categorical distribution using \texttt{numpy.random.Generator.choice}, ensuring that the simulated cohort reproduces the intended socioeconomic structure in expectation. This step constitutes the root node of the data-assembly pipeline, as downstream behavioural (\textit{e.g.,} smoking), anthropometric (\textit{e.g.,} BMI), and clinical (\textit{e.g.,} CKD) variables are subsequently drawn conditional on these IRSD quintiles.

\begin{lstlisting}[language=Python, style=mystyle, backgroundcolor=\color{backcolour4Input}]
IRSD_WEIGHTS = {1: 0.2136, 2: 0.1622, 3: 0.2393, 4: 0.1678, 5: 0.2171}

IRSD_QUINTILES = np.array([1, 2, 3, 4, 5])
IRSD_PROBS     = np.array([IRSD_WEIGHTS[q] for q in IRSD_QUINTILES])

def sample_irsd_quintile(n, rng=None):
    if rng is None:
        rng = np.random.default_rng(RNG_SEED)
    return rng.choice(IRSD_QUINTILES, size=n, p=IRSD_PROBS)
\end{lstlisting}

\subsection{Age Sampling}\label{App:A13}
We generate individual ages by sampling from a Gaussian distribution calibrated to empirical primary-prevention cohort characteristics, with mean $49.80$ years and standard deviation $12.39$. Since real-world cardiovascular prevention studies typically impose eligibility constraints and because chronological age is naturally bounded, we apply a deterministic truncation step to confine sampled ages to the interval $[18,90]$. This clipping acts as a simple but effective approximation to a truncated normal without requiring rejection sampling. The lower bound ensures exclusion of paediatric profiles, while the upper bound prevents implausible outliers that would distort downstream risk-factor distributions or hazard calibration. The resulting age vector serves as one of the principal exogenous covariates in the data-assembly pipeline, influencing the conditional models for diabetes, CKD, SBP, eGFR, and atrial fibrillation. All draws use a user-supplied or default pseudorandom number generator to maintain reproducibility within the full synthetic-cohort pipeline.

\begin{lstlisting}[language=Python, style=mystyle, backgroundcolor=\color{backcolour4Input}]
AGE_MEAN = 49.80
AGE_SD   = 12.39
AGE_MIN  = 18
AGE_MAX  = 90

def sample_age(n, rng=None):
    if rng is None:
        rng = np.random.default_rng(RNG_SEED)
    age = rng.normal(loc=AGE_MEAN, scale=AGE_SD, size=n)
    return np.clip(age, AGE_MIN, AGE_MAX)
\end{lstlisting}

\subsection{Smoking Status Conditional on IRSD}\label{App:A14}
To encode socioeconomic patterning in health behaviours, smoking status is generated as a categorical variable whose distribution is conditioned explicitly on each individual’s assigned IRSD quintile. For each quintile $q\in{1,\ldots,5}$, we specify a triplet of probabilities governing the likelihood of being a non-smoker, current smoker, or ex-smoker, with lower socioeconomic strata (higher deprivation) exhibiting higher rates of current smoking in accordance with population surveillance statistics. Given a vector of IRSD assignments, we partition the cohort into quintile-specific subsets and, for each subset, independently draw smoking statuses using \texttt{numpy.random.Generator.choice} with the quintile-appropriate probability vector $(p_{non}, p_{current}, p_{ex})$. This stratified sampling procedure preserves the conditional distribution $\mathbb{P}(\text{smoking}\vert\text{IRSD})$ exactly in expectation and operationalises the causal edge IRSD $\rightarrow$ smoking in our data-generating DAG. As downstream cardiometabolic and clinical variables (\textit{e.g.,} CKD, SBP, AF) incorporate smoking as an input, this step ensures that socioeconomic gradients propagate appropriately through the assembly cohort.

\begin{lstlisting}[language=Python, style=mystyle, backgroundcolor=\color{backcolour4Input}]
SMOKING_STATES = np.array(["non", "current", "ex"])

SMOKING_PROB_MAP_IRSD = {
    1: {"non": 0.64,  "current": 0.16,  "ex": 0.20},
    2: {"non": 0.695, "current": 0.125, "ex": 0.18},
    3: {"non": 0.73,  "current": 0.10,  "ex": 0.17},
    4: {"non": 0.775, "current": 0.075, "ex": 0.15},
    5: {"non": 0.81,  "current": 0.05,  "ex": 0.14},
}

def sample_smoking_given_irsd(irsd_quintile, rng=None):
    if rng is None:
        rng = np.random.default_rng(RNG_SEED)
    n = len(irsd_quintile)
    out = np.empty(n, dtype=object)
    for q in IRSD_QUINTILES:
        idx = np.where(irsd_quintile == q)[0]
        if len(idx) == 0:
            continue
        p = SMOKING_PROB_MAP_IRSD[q]
        probs = np.array([p["non"], p["current"], p["ex"]])
        out[idx] = rng.choice(SMOKING_STATES, size=len(idx), p=probs)
    return out
\end{lstlisting}

\subsection{BMI Conditional on IRSD}\label{App:A15}
To encode socioeconomic gradients in adiposity, we model BMI as a continuous variable whose mean shifts systematically across IRSD quintiles. Specifically, we construct a quintile-specific mean function $\mu_q$ that decreases linearly with socioeconomic advantage via a slope parameter $\Delta_{\mathrm{BMI}} = 0.6$, centred on the middle quintile ($q = 3$). Since this linear shift must remain consistent with the empirically observed population-wide BMI mean ($28.29$), we first solve for a central offset $\mu_{\mathrm{center}}$ such that the weighted average $\sum_q w_q \mu_q$ (using the IRSD population weights $w_q$) reproduces the global target exactly; this ensures demographic realism even when assembly cohorts depart from the original IRSD distribution. Given these calibrated means, BMI values are drawn independently within each IRSD stratum from Gaussian distributions with standard deviation $4.98$, after which values are truncated to the physiologically plausible range $[15, 60]$. This construction explicitly encodes the causal relationship IRSD~$\rightarrow$~BMI in the data-generating DAG and propagates socioeconomic influences into downstream metabolic and clinical processes (\textit{e.g.,} diabetes, CKD, SBP).

\begin{lstlisting}[language=Python, style=mystyle, backgroundcolor=\color{backcolour4Input}]
BMI_MEAN = 28.29
BMI_SD   = 4.98
BMI_MIN  = 15.0
BMI_MAX  = 60.0
DELTA_BMI = 0.6
Q_CENTER  = 3

def compute_bmi_means_by_irsd():
    w = IRSD_WEIGHTS
    S = sum(w[q] * (Q_CENTER - q) for q in IRSD_QUINTILES)
    mu_center = BMI_MEAN - DELTA_BMI * S
    mu = {}
    for q in IRSD_QUINTILES:
        mu[q] = mu_center + DELTA_BMI * (Q_CENTER - q)
    return mu

BMI_MEANS_BY_IRSD = compute_bmi_means_by_irsd()

def sample_bmi_given_irsd(irsd_quintile, rng=None):
    if rng is None:
        rng = np.random.default_rng(RNG_SEED)
    n = len(irsd_quintile)
    bmi = np.empty(n, dtype=float)
    for q in IRSD_QUINTILES:
        idx = np.where(irsd_quintile == q)[0]
        if len(idx) == 0:
            continue
        mean_q = BMI_MEANS_BY_IRSD[q]
        bmi[idx] = rng.normal(loc=mean_q, scale=BMI_SD, size=len(idx))
    return np.clip(bmi, BMI_MIN, BMI_MAX)
\end{lstlisting}

\subsection{Diabetes Conditional on Age and BMI}\label{App:A16}
We model the probability of baseline diabetes using a logistic regression calibrated to reproduce both the marginal prevalence and established epidemiological risk gradients. Let \text{Age} and \text{BMI} denote the sampled covariates, centred at their population means. The log-odds of diabetes are parameterised as
\[
\text{logit}\, P(\mathrm{DM}=1 \mid \text{Age}, \text{BMI})
=
\alpha_{\mathrm{DM}}
\;+\;
\beta_{\mathrm{Age}}
\left( \frac{\text{Age} - \text{AGE\_MEAN}}{10} \right)
\;+\;
\beta_{\mathrm{BMI}}
(\text{BMI} - \text{BMI\_MEAN}),
\]
where
\[
\alpha_{\mathrm{DM}}
=
\log\!\left(
\frac{0.0553}{1 - 0.0553}
\right)
\]
encodes a baseline prevalence of 5.53\% at mean age and BMI. The age coefficient
\[
\beta_{\mathrm{Age}} = \log(1.8)
\]
reflects an odds ratio of 1.8 per 10-year increase, while
\[
\beta_{\mathrm{BMI}} = \frac{\log(1.7)}{5}
\]
encodes an odds ratio of 1.7 per 5-unit increase in BMI. These effect sizes approximate population-level associations observed in large epidemiological cohorts and ensure realistic age--BMI--diabetes interactions within the assembly cohort. For each individual, we compute the corresponding probability $p$ via the logistic function and draw a Bernoulli outcome to obtain a binary diabetes indicator. This construction operationalises the causal edges $\text{Age} \rightarrow \mathrm{DM}$ and $\text{BMI} \rightarrow \mathrm{DM}$ in the baseline data-generating DAG.

\begin{lstlisting}[language=Python, style=mystyle, backgroundcolor=\color{backcolour4Input}]
BASE_DM_PREV   = 0.0553
OR10_AGE_DM    = 1.8
OR5_BMI_DM     = 1.7

BETA_AGE_DM = np.log(OR10_AGE_DM)
BETA_BMI_DM = np.log(OR5_BMI_DM) / 5.0

ALPHA_DM = np.log(BASE_DM_PREV / (1 - BASE_DM_PREV))

def sample_diabetes_given_age_bmi(age, bmi, rng=None):
    if rng is None:
        rng = np.random.default_rng(RNG_SEED)
    age_scaled   = (age - AGE_MEAN) / 10.0
    bmi_centered = (bmi - BMI_MEAN)
    lin = ALPHA_DM + BETA_AGE_DM * age_scaled + BETA_BMI_DM * bmi_centered
    p   = logistic(lin)
    return rng.binomial(1, p, size=len(age))
\end{lstlisting}

\subsection{CKD Conditional on Age, BMI, Diabetes, and Smoking}\label{App:A17}
We model CKD as a binary clinical condition generated via a logistic regression whose parameters are calibrated to reflect both its low population prevalence and its established risk-factor associations. The baseline log-odds term
\[
\alpha_{\mathrm{CKD}}
=
\log\!\left(
\frac{0.0045}{1 - 0.0045}
\right)
\]
encodes a target prevalence of 0.45\% at mean age and BMI, in the absence of diabetes and for non-smokers. Deviations from these reference covariate values enter additively on the log-odds scale: age increases CKD odds with coefficient
\[
\beta_{\mathrm{Age}} = \frac{\log(1.4)}{10},
\]
corresponding to an odds ratio of 1.4 per 10-year increase; elevated BMI raises CKD risk with coefficient
\[
\beta_{\mathrm{BMI}} = \frac{\log(1.5)}{5};
\]
diabetes confers a threefold increase in odds,
\[
\beta_{\mathrm{DM}} = \log(3.0);
\]
and smoking status contributes additional shifts, with ex-smokers and current smokers receiving log-odds increments
\[
\beta_{\mathrm{ex}} = \log(1.2)
\qquad\text{and}\qquad
\beta_{\mathrm{curr}} = \log(1.4),
\]
respectively. For each individual, we compute the linear predictor by centering age and BMI at their population means, add the smoking- and diabetes-specific adjustments, transform the result through the logistic function to obtain a probability $p$, and finally draw a Bernoulli outcome. This module encodes the causal pathways $\text{Age} \rightarrow \mathrm{CKD}$, $\text{BMI} \rightarrow \mathrm{CKD}$, $\mathrm{DM} \rightarrow \mathrm{CKD}$, and $\text{Smoking} \rightarrow \mathrm{CKD}$ in the cohort assembly-generating DAG, thereby propagating behavioural and metabolic risk into renal impairment.

\begin{lstlisting}[language=Python, style=mystyle, backgroundcolor=\color{backcolour4Input}]
TARGET_CKD_PREV      = 0.0045
OR10_AGE_CKD         = 1.4
OR_DM_CKD            = 3.0
OR5_BMI_CKD          = 1.5
OR_EX_VS_NON_CKD     = 1.2
OR_CURR_VS_NON_CKD   = 1.4

BETA_AGE_CKD    = np.log(OR10_AGE_CKD) / 10.0
BETA_DM_CKD     = np.log(OR_DM_CKD)
BETA_BMI_CKD    = np.log(OR5_BMI_CKD) / 5.0
BETA_EX_CKD     = np.log(OR_EX_VS_NON_CKD)
BETA_CURR_CKD   = np.log(OR_CURR_VS_NON_CKD)

ALPHA_CKD = np.log(TARGET_CKD_PREV / (1 - TARGET_CKD_PREV))

def sample_ckd(age, bmi, diabetes, smoking_status, rng=None):
    if rng is None:
        rng = np.random.default_rng(RNG_SEED)

    age_centered = age - AGE_MEAN
    bmi_centered = bmi - BMI_MEAN

    lin = ALPHA_CKD
    lin += BETA_AGE_CKD * age_centered
    lin += BETA_DM_CKD * diabetes
    lin += BETA_BMI_CKD * bmi_centered
    lin[smoking_status == "ex"]      += BETA_EX_CKD
    lin[smoking_status == "current"] += BETA_CURR_CKD

    p = logistic(lin)
    return rng.binomial(1, p, size=len(age))
\end{lstlisting}

\subsection{HbA1c Conditional on Diabetes}\label{App:A18}
In order to assemble glycaemic biomarker values that reflect clinically realistic heterogeneity, we model HbA1c using a two-component Gaussian mixture distribution conditioned on diabetes status. Individuals without diabetes draw their HbA1c values from a normal distribution with mean $4.60$ and standard deviation $0.60$, representing normoglycaemic physiology with relatively low variability. In contrast, individuals with diabetes draw from a broader distribution with mean $7.10$ and standard deviation $1.20$, capturing both elevated glycaemic burden and increased inter-individual dispersion commonly observed in treated and untreated diabetic populations. The model therefore produces a clear separation between diabetic and non-diabetic groups while still allowing physiologically plausible overlap. This conditional structure operationalizes the causal edge $\mathrm{DM} \rightarrow \mathrm{HbA1c}$ in the cohort assembling DAG and ensures that diabetes---rather than downstream variables---serves as the primary determinant of glycaemic elevation in the synthetic cohort.

\begin{lstlisting}[language=Python, style=mystyle, backgroundcolor=\color{backcolour4Input}]
HBA1C_MEAN_NODM = 4.60
HBA1C_SD_NODM   = 0.60
HBA1C_MEAN_DM   = 7.10
HBA1C_SD_DM     = 1.20

def sample_hba1c(diabetes, rng=None):
    if rng is None:
        rng = np.random.default_rng(RNG_SEED)
    diabetes = np.asarray(diabetes)
    mean = np.where(diabetes == 1, HBA1C_MEAN_DM, HBA1C_MEAN_NODM)
    sd   = np.where(diabetes == 1, HBA1C_SD_DM,   HBA1C_SD_NODM)
    return rng.normal(loc=mean, scale=sd, size=len(diabetes))
\end{lstlisting}

\subsection{eGFR Conditional on Age and CKD}\label{App:A19}
eGFR is modelled as a continuous physiological marker whose mean varies systematically with age and CKD status, reflecting clinically established patterns of renal function decline. For an individual of age~\text{Age} and CKD indicator $\text{CKD} \in \{0,1\}$, we define the expected eGFR as
\[
\mu
=
\mathrm{EGFR\_MEAN}
+
\beta_{\mathrm{Age}}
\,(\text{Age} - \mathrm{AGE\_MEAN})
+
\Delta_{\mathrm{CKD}} \cdot \text{CKD},
\]
where the age slope $\beta_{\mathrm{Age}} = -0.18$ encodes an average decline of $0.18$ mL/min/1.73m$^2$ per year relative to the population mean, and the CKD shift $\Delta_{\mathrm{CKD}} = -31.0$ imposes a large downward offset consistent with clinical reductions in filtration capacity among individuals with diagnosed CKD. To capture residual biological and measurement variability, we draw eGFR values from a normal distribution with standard deviation $5.0$ around this mean. This generative mechanism encodes the causal relationships $\text{Age} \rightarrow \text{eGFR}$ and $\text{CKD} \rightarrow \text{eGFR}$ in the baseline data-generating DAG and ensures that renal function declines in a physiologically plausible manner across the assembled cohort.

\begin{lstlisting}[language=Python, style=mystyle, backgroundcolor=\color{backcolour4Input}]
EGFR_MEAN = 82.97
BETA_AGE_EGFR = -0.18  # per year (approx -1.8 per 10 yrs)
EGFR_SD_RESID = 5.0
DELTA_EGFR_CKD = -31.0  # CKD=1 vs 0

def sample_egfr(age, ckd, rng=None):
    if rng is None:
        rng = np.random.default_rng(RNG_SEED)
    age = np.asarray(age)
    ckd = np.asarray(ckd)
    mean = EGFR_MEAN + BETA_AGE_EGFR * (age - AGE_MEAN) + DELTA_EGFR_CKD * ckd
    return rng.normal(loc=mean, scale=EGFR_SD_RESID, size=len(age))
\end{lstlisting}

\subsection{SBP Conditional on Age, BMI, Diabetes, CKD, and Smoking}\label{App:A20}
SBP is assembled using a linear mean model that integrates demographic, metabolic, renal, and behavioural determinants known to influence hypertension risk. For an individual with age~\text{Age}, BMI~\text{BMI}, diabetes status~$\mathrm{DM}$, CKD status~$\mathrm{CKD}$, and categorical smoking behaviour, we define the expected SBP as
\begin{align*}
\mu_{\mathrm{SBP}}
&=
\mathrm{SBP\_MEAN\_TARGET}
+
\beta_{\mathrm{Age}} \, (\text{Age} - \mathrm{AGE\_MEAN})
+
\beta_{\mathrm{BMI}} \, (\text{BMI} - \mathrm{BMI\_MEAN})
\\
&\quad
+
\Delta_{\mathrm{DM}} \cdot \mathrm{DM}
+
\Delta_{\mathrm{CKD}} \cdot \mathrm{CKD}
+
\Delta_{\mathrm{smoke}}(\text{smoking}),
\end{align*}
where $\beta_{\mathrm{Age}} = 0.38$ and $\beta_{\mathrm{BMI}} = 1.0$ encode positive age- and adiposity-related increases in SBP, while diabetes and CKD contribute fixed upward shifts of $8.0$ and $12.0$~mmHg, respectively, consistent with clinical patterns of cardiometabolic and renal hypertension. Smoking introduces an additional categorical shift, with ex-smokers receiving a $+1.7$~mmHg adjustment and current smokers $+3.8$~mmHg. Around this mean, we draw SBP values from a normal distribution with residual standard deviation $14.0$ to reflect biological variability and measurement noise. This specification explicitly encodes the causal pathways $\text{Age}, \text{BMI}, \mathrm{DM}, \mathrm{CKD}, \text{Smoking} \rightarrow \mathrm{SBP}$ in the baseline cohort-assembling DAG, ensuring realistic joint correlations among cardiometabolic risk factors.

\begin{lstlisting}[language=Python, style=mystyle, backgroundcolor=\color{backcolour4Input}]
SBP_MEAN_TARGET = 122.07
BETA_AGE_SBP    = 0.38
BETA_BMI_SBP    = 1.0
DELTA_SBP_EX      = 1.7
DELTA_SBP_CURRENT = 3.8
DELTA_SBP_DM      = 8.0
DELTA_SBP_CKD     = 12.0

SBP_RESID_SD = 14.0

def sample_sbp(age, bmi, diabetes, ckd, smoking_status, rng=None):
    if rng is None:
        rng = np.random.default_rng(RNG_SEED)

    shift_smoke = np.zeros(len(age))
    shift_smoke[smoking_status == "ex"]      = DELTA_SBP_EX
    shift_smoke[smoking_status == "current"] = DELTA_SBP_CURRENT

    mean_sbp = (
        SBP_MEAN_TARGET
        + BETA_AGE_SBP * (age - AGE_MEAN)
        + BETA_BMI_SBP * (bmi - BMI_MEAN)
        + DELTA_SBP_DM * diabetes
        + DELTA_SBP_CKD * ckd
        + shift_smoke
    )
    return rng.normal(loc=mean_sbp, scale=SBP_RESID_SD, size=len(age))
\end{lstlisting}

\subsection{Atrial Fibrillation Conditional on Age, CKD, and Smoking}\label{App:A21}
AF is assembled using a logistic regression model calibrated to match both its low marginal prevalence and its canonical epidemiological risk gradients. The baseline log-odds term
\[
\alpha_{\mathrm{AF}}
=
\log\!\left(
\frac{0.0057}{1 - 0.0057}
\right)
\]
corresponds to an AF prevalence of 0.57\% at mean age, without CKD, and among non-smokers. Age-related risk is introduced through a coefficient
\[
\beta_{\mathrm{Age}} = \frac{\log(1.7)}{10},
\]
encoding an odds ratio of 1.7 per 10-year increase. CKD strongly raises AF risk, contributing an additive log-odds shift
\[
\beta_{\mathrm{CKD}} = \log(3.0),
\]
while smoking adds behaviour-specific increments, with ex-smokers receiving
\[
\beta_{\mathrm{ex}} = \log(1.1)
\quad\text{and current smokers}\quad
\beta_{\mathrm{curr}} = \log(1.3).
\]
For each individual, we construct the linear predictor by centering age at its population mean and adding CKD- and smoking-specific effects before applying the logistic function to obtain a probability $p$. A Bernoulli draw yields the AF indicator. This model therefore encodes the causal effects $\text{Age} \rightarrow \mathrm{AF}$, $\mathrm{CKD} \rightarrow \mathrm{AF}$, and $\text{Smoking} \rightarrow \mathrm{AF}$ within the baseline cohort-assembling DAG, ensuring coherent propagation of cardiac and renal risk factors.

\begin{lstlisting}[language=Python, style=mystyle, backgroundcolor=\color{backcolour4Input}]
TARGET_AF_PREV      = 0.0057
OR10_AGE_AF         = 1.7
OR_CKD_AF           = 3.0
OR_EX_VS_NON_AF     = 1.1
OR_CURR_VS_NON_AF   = 1.3

BETA_AGE_AF   = np.log(OR10_AGE_AF) / 10.0
BETA_CKD_AF   = np.log(OR_CKD_AF)
BETA_EX_AF    = np.log(OR_EX_VS_NON_AF)
BETA_CURR_AF  = np.log(OR_CURR_VS_NON_AF)

ALPHA_AF = np.log(TARGET_AF_PREV / (1 - TARGET_AF_PREV))

def sample_af(age, ckd, smoking_status, rng=None):
    if rng is None:
        rng = np.random.default_rng(RNG_SEED)

    age = np.asarray(age)
    ckd = np.asarray(ckd)
    smoking_status = np.asarray(smoking_status)

    age_centered = age - AGE_MEAN

    lin = ALPHA_AF + BETA_AGE_AF * age_centered + BETA_CKD_AF * ckd
    lin[smoking_status == "ex"]      += BETA_EX_AF
    lin[smoking_status == "current"] += BETA_CURR_AF

    p = logistic(lin)
    return rng.binomial(1, p, size=len(age))
\end{lstlisting}

%===========================================================
%===========================================================
%===========================================================
\newpage
\section{Relational Asset Construction (Data Asset 2 Implementation)}\label{App:B01}

This appendix documents the precise implementation steps used to transform PRIME-CVD Data Asset 1 into the EMR-style relational dataset Data Asset 2. Three tables are constructed:\\
\hspace*{5mm}\texttt{[PatientEMR].[PatientMasterSummary]},\\
\hspace*{5mm}\texttt{[PatientEMR].[PatientChronicDiseases]}, and\\ 
\hspace*{5mm}\texttt{[PatientEMR].[PatientMeasAndPath]}.\\
These tables collectively emulate real-world EMR artefacts including non-sequential identifiers, missingness, heterogeneous terminology, and unit inconsistencies.

\begin{figure}[h!]
    \centering
    \includegraphics[width=\linewidth]{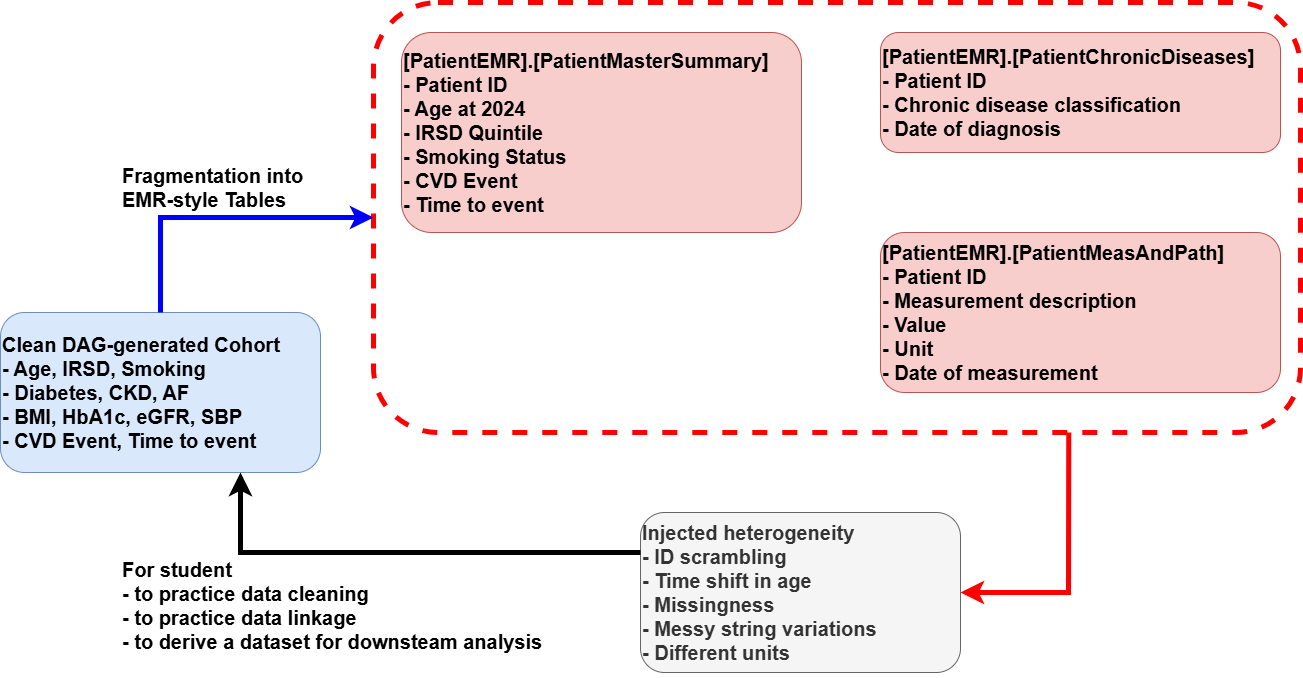}
    \caption{\newline Constructing the PRIME-CVD EMR-style data asset.\newline
    \hspace*{ 5mm} The clean DAG-generated cohort is \newline
    \hspace*{10mm} split into three relational tables (red) and \newline 
    \hspace*{10mm} augmented with realistic EMR artefacts (grey), including\newline
    \hspace*{15mm} missingness,\newline
    \hspace*{15mm} ID\newline 
    \hspace*{15mm} scrambling,\newline 
    \hspace*{15mm} heterogeneous terminology, and\newline 
    \hspace*{15mm} mixed units\newline
    \hspace*{10mm} to create Data Asset 2.}
\end{figure}

%===========================================================
% MASTER SUMMARY CONSTRUCTION
%===========================================================
\newpage
\subsection{Construction of \texttt{[PatientEMR].[PatientMasterSummary]}}\label{App:B11}

The \texttt{[PatientMasterSummary]} table constitutes the first stage in transforming the clean DAG-generated cohort into an EMR-style relational structure. Each individual in the clean cohort is assigned a synthetic identifier constructed deterministically from the row index (via a non-linear offset-and-scaling transformation) to mirror patient IDs that bear no transparent relationship to underlying data ordering while remaining reproducible within the simulation. Age is reframed as ``Age at~2024'' by adding seven years to the baseline age, reflecting a common EMR pattern in which stored ages correspond to a fixed extraction year rather than biological age at risk-factor assessment. Smoking status is carried forward from the clean cohort but is degraded by injecting missingness in 15.66\% of non-smokers, approximating real-world primary-care under-documentation of negative lifestyle factors. IRSD quintile is preserved without modification, as socioeconomic status is typically stable and well-coded in administrative datasets. Cardiovascular outcomes are encoded in two components: a binary event indicator and a coarse-grained year--month timestamp. Event times derive from the simulated continuous follow-up time, mapped to a calendar scale with origin 2017--01--01 and formatted as \texttt{YYYY--MM} to reflect EMR date resolution; censored individuals are assigned a terminal month (``2022--12'') corresponding to the extraction boundary. This construction produces a patient-level summary table that preserves essential risk factors while embodying characteristic EMR artifacts -- identifier scrambling, age shifting, lifestyle missingness, discrete date recording, and separation between event occurrence and censoring logic.

\begin{lstlisting}[language=Python, style=mystyle, backgroundcolor=\color{backcolour4Input}]
df_emr = pd.DataFrame()

# Synthetic patient ID and age at 2024
X = df0.index.astype(int)
df_emr["Patient_ID"]  = (X**2 - 77) * 3 + 500
df_emr["Age_At_2024"] = np.round(df0["Age"] + 7, 2)

# Socioeconomic position and smoking (with injected missingness)
df_emr["IRSD_Quintile"]  = df0["IRSD_quintile"]
df_emr["SMOKING_STATUS"] = df0["smoking_status"]
mask_non   = df_emr["SMOKING_STATUS"].eq("non")
idx_to_nan = df_emr[mask_non].\
             sample(frac=0.1566, random_state=42).index
df_emr.loc[idx_to_nan, "SMOKING_STATUS"] = np.nan

# Cardiovascular event indicator and coarse event/censoring time
df_emr["CVD_Event"] = df0["cvd_event"].astype(int)
base_date   = pd.to_datetime("2017-01-01")
event_dates = base_date +\ 
              pd.to_timedelta(df0["cvd_time"] * 365.25, unit="D")
df_emr["CVD_Time"]  = event_dates.dt.strftime("%Y-%m")
df_emr.loc[df_emr["CVD_Event"] == 0, "CVD_Time"] = "2022-12"
\end{lstlisting}

%===========================================================
% CHRONIC DISEASES CONSTRUCTION
%===========================================================
%\newpage
\subsection{Construction of \texttt{[PatientEMR].[PatientChronicDiseases]}}\label{App:B12}

The \texttt{[PatientChronicDiseases]} table transforms the clean cohort’s binary disease indicators (diabetes, CKD, and AF) into a long-form EMR-style diagnosis record for each affected individual. Each diagnosis entry inherits the synthetic patient identifier and is assigned a diagnosis month sampled uniformly between 2012--01 and 2016--12, reflecting typical temporal dispersion of historical problem-list entries in primary-care systems. To emulate real-world coding heterogeneity, each disease label is replaced by a randomly drawn term from a curated vocabulary of clinically plausible synonyms, abbreviations, and ICD9/ICD10 codes, with probabilities chosen to mimic their relative prevalence in legacy EMRs (\textit{e.g.,} ``Diabetes'', ``T2DM'', ``ICD10:~E11''). This construction yields a diagnosis table in which conditions appear multiple times across heterogeneous string forms, requiring downstream harmonisation during cohort reconstruction. The resulting table thus embodies three characteristic EMR artefacts: one-to-many expansion of conditions, heterogeneous terminology, and non-informative timestamp granularity.

\begin{lstlisting}[language=Python, style=mystyle, backgroundcolor=\color{backcolour4Input}]
# Identify patients with diabetes, CKD, and AF
flags = df0[["diabetes", "CKD", "AF"]].astype(int)
flags["Patient_ID"] = (df0.index**2 - 77) * 3 + 500

# Long-form expansion: one row per patient-condition
df_dm  = flags.loc[flags["diabetes"] == 1, ["Patient_ID"]].\
         assign(Category="Diabetes")
df_ckd = flags.loc[flags["CKD"]      == 1, ["Patient_ID"]].\
         assign(Category="Chronic Kidney Disease")
df_af  = flags.loc[flags["AF"]       == 1, ["Patient_ID"]].\
         assign(Category="Atrial Fibrillation")
df_flags = pd.concat([df_dm, df_ckd, df_af], ignore_index=True)

# Assign diagnosis months (2012-2016)
months = pd.date_range("2012-01-01", "2016-12-01", freq="MS")
df_flags["Date"] = np.random.default_rng(42).\
                   choice(months, len(df_flags)).\
                   astype("datetime64[M]")

# Replace canonical labels with heterogeneous EMR-style terminology
df_flags.loc[df_flags["Category"] == "Diabetes", 
             "Category"] =\
    np.random.default_rng(123).\
    choice(choices_dm, p=probs_dm, 
           size=(df_flags["Category"] == "Diabetes").sum())

df_flags.loc[df_flags["Category"] == "Atrial Fibrillation", 
             "Category"] =\
    np.random.default_rng(123).\
    choice(choices_af, p=probs_af, 
           size=(df_flags["Category"] == "Atrial Fibrillation").sum())

df_flags.loc[df_flags["Category"] == "Chronic Kidney Disease",
             "Category"] =\ 
    np.random.default_rng(123).\
    choice(choices_ckd, p=probs_ckd, 
    size=(df_flags["Category"] == "Chronic Kidney Disease").sum())

# Shuffle rows to obscure ordering
df_flags = df_flags.sample(frac=1.0, random_state=999).\
           reset_index(drop=True)
\end{lstlisting}

%===========================================================
% MEASUREMENTS & PATHOLOGY CONSTRUCTION
%===========================================================
%\newpage
\subsection{Construction of \texttt{[PatientEMR].[PatientMeasAndPath]}}\label{App:B13}

The \texttt{[PatientMeasAndPath]} table converts the clean cohort’s baseline biomarkers into a long-form EMR-style laboratory and vital-signs record. For each individual, we create separate rows for HbA1c, eGFR, and SBP, each linked to the synthetic patient identifier and the corresponding numeric value from the DAG-generated cohort. Every measurement is assigned a pseudo-observation month sampled uniformly between 2012--01 and 2016--12, reflecting the diffuse timing of historical test results in routine primary care. To emulate real-world heterogeneity in test naming, we replace the canonical measure labels with strings drawn from curated vocabularies of synonymous or near-synonymous terms (including common abbreviations, spelling variants, and LOINC-style codes), with probability weights chosen to approximate their relative prevalence in legacy systems. Units are assigned consistently for each measure (mmHg for SBP, mL/min/1.73m$^2$ for eGFR, and \% for HbA1c), after which a random 5\% subsample of HbA1c results is converted from percent to IFCC mmol/mol units using the standard linear transformation and relabelled accordingly. Finally, the table is row-shuffled to obscure any residual ordering structure. This design yields a measurement table that is faithful to the underlying biology yet exhibits the string-level and unit-level inconsistencies that analysts must routinely reconcile when building analysis-ready cohorts from EMR data.

\begin{lstlisting}[language=Python, style=mystyle, backgroundcolor=\color{backcolour4Input}]
# Long-form biomarker table: one row per patient-measure
pid = (df0.index**2 - 77) * 3 + 500
df_hba1c = pd.DataFrame({"Patient_ID": pid, 
                         "Measure": "HbA1c", "Value": df0["HbA1c"]})
df_egfr  = pd.DataFrame({"Patient_ID": pid, 
                         "Measure": "eGFR",  "Value": df0["eGFR"]})
df_sbp   = pd.DataFrame({"Patient_ID": pid, 
                         "Measure": "SBP",   "Value": df0["SBP"]})
baseline_biomarkers = pd.concat([df_hba1c, df_egfr, df_sbp], 
                                ignore_index=True)

# Assign observation months (2012-2016)
months = pd.date_range("2012-01-01", "2016-12-01", freq="MS")
rng_dates = np.random.default_rng(42)
baseline_biomarkers["Date"] =\
    rng_dates.choice(months, len(baseline_biomarkers)).\
    astype("datetime64[M]")

# Heterogeneous EMR-style descriptions
# (choices_* and probs_* defined elsewhere)
rng_terms = np.random.default_rng(123)
mask_hba1c = baseline_biomarkers["Measure"].eq("HbA1c")
mask_egfr  = baseline_biomarkers["Measure"].eq("eGFR")
mask_sbp   = baseline_biomarkers["Measure"].eq("SBP")

baseline_biomarkers.loc[mask_hba1c, "Description"] =\
    rng_terms.choice(
        choices_hba1c, size=mask_hba1c.sum(), p=probs_hba1c)
baseline_biomarkers.loc[mask_egfr,  "Description"] =\
    rng_terms.choice(
        choices_egfr, size=mask_egfr.sum(), p=probs_egfr)
baseline_biomarkers.loc[mask_sbp,   "Description"] =\ 
    rng_terms.choice(
        choices_sbp, size=mask_sbp.sum(), p=probs_sbp)

# Assign units and convert 5% of HbA1c results to mmol/mol
baseline_biomarkers["Unit"] = None
baseline_biomarkers.loc[mask_sbp,   "Unit"] = "mmHg"
baseline_biomarkers.loc[mask_egfr,  "Unit"] = "mL/min/1.73m2"
baseline_biomarkers.loc[mask_hba1c, "Unit"] = "%"

rng_conv = np.random.default_rng(2025)
hba_idx  = baseline_biomarkers.index[mask_hba1c]
convert_idx = rng_conv.choice(
                hba_idx, size=int(0.05 * len(hba_lidx)), replace=False)
vals_pct = baseline_biomarkers.loc[convert_idx, "Value"].astype(float)
baseline_biomarkers.loc[convert_idx, "Value"] =\
    (vals_pct - 2.15) * 10.929
baseline_biomarkers.loc[convert_idx, "Unit"]  =\
    "mmol/mol"

# Shuffle rows to obscure ordering
baseline_biomarkers =\
    baseline_biomarkers.\
    sample(frac=1.0, random_state=999).\
    reset_index(drop=True)
\end{lstlisting}

%===========================================================
%===========================================================
%===========================================================
\newpage
\section{Injected Messiness in Data Asset 2}\label{App:C01}

We inject heterogeneous terminology, inconsistent units and lexical noise into Data Asset 2.

\begin{table}[htbp]
\centering
\footnotesize
\renewcommand{\arraystretch}{1.25}
\caption{Sampling distributions governing injected messiness in the relational data asset.}
\begin{tabular}{p{1.5cm} p{3cm} p{8cm}}
\hline
\textbf{Target} & \textbf{Sampling Rules} & \textbf{Description of Injected Heterogeneity} \\
\hline

\textbf{Smoking Status\newline Missingness} 
& Bernoulli($p = 0.1566$) applied to non-smokers 
& 15.66\% of patients coded as ``non'' have their value set to \texttt{NaN}, producing patterned lifestyle missingness consistent with real GP EMRs. No missingness is applied to ``ex'' or ``current'' smokers. \\\\

\hline

\textbf{Diabetes Terms} 
& Multinomial over\newline 6 labels  
\newline $[0.40, 0.20, 0.15,$\newline $ 0.15, 0.05, 0.05]$
& Six heterogeneous term (\texttt{``Diabetes''}, \texttt{``T2DM''}, \texttt{ICD10/ICD9 strings}, lay terms) are sampled for each diabetic patient.\newline Ensures realistic variation in coding systems across time and clinicians. \\\\

\hline

\textbf{AF\newline Terms} 
& Multinomial over\newline 6 labels  
\newline $[0.45, 0.30, 0.075,$\newline $ 0.075, 0.05, 0.05]$
& Blends abbreviations, long-form names, and ICD codes (\textit{e.g.,} \texttt{``AF''}, \texttt{``Atrial fibrillation''}, \texttt{``A-fib''}) to generate lexical diversity typical of EMR problem lists. \\\\

\hline

\textbf{CKD\newline Terms} 
& Multinomial over\newline 7 labels  
\newline $[0.45, 0.25, 0.05,$\newline $ 0.05, 0.05, 0.10, 0.05]$
& Produces a mixture of short labels, verbose forms, renal-function descriptors, and ICD codes (\textit{e.g.,} \texttt{``CKD''}, \texttt{``Chronic renal failure''}, \texttt{``Renal insufficiency''}). \\\\

\hline

\textbf{HbA1c\newline Labels} 
& Multinomial over\newline 10 labels  
\newline dominant class: 55\%
& Sampling includes canonical labels (55\%), upper-case variants (15\%), abbreviations (10\%), verbose names (5\%), and rare LOINC-like coding (4\%). Provides realistic heterogeneity requiring text harmonisation. \\\\

\hline

\textbf{HbA1c Unit\newline Inconsistency} 
& Bernoulli($p = 0.05$) on HbA1c rows 
& 5\% of HbA1c values are converted from \% to mmol/mol using the IFCC formula. Produces unit heterogeneity requiring correct back-transformation during cleaning. \\\\

\hline

\textbf{eGFR \newline Labels} 
& Multinomial over\newline 10 labels  
\newline dominant class: 60\%
& Labels range from canonical (\texttt{``eGFR''}) to capitalised forms, verbose descriptions, LOINC-style pseudo-codes, and typographical variants (~0.3\%). Encourages robust mapping strategies. \\\\

\hline

\textbf{SBP \newline Labels} 
& Multinomial over\newline 12 labels  
\newline dominant class: 50\%
& Includes \texttt{``SBP''}, expanded terms, multiple spacing/case variants, and rare (~0.7\%) typographic noise, mirroring inconsistencies in vital-sign extraction from historic systems. \\\\

\hline

\textbf{Measurement Dates} 
& Uniform distribution\newline over\newline 
2012–01 to 2016–12\newline (monthly)
& Provides temporally diffuse historical test dates independent of baseline cohort timing. Breaks alignment between sampling, diagnosis, and outcome times, similar to real EMR backfill. \\\\

\hline

\textbf{Diagnosis Dates} 
& Uniform distribution\newline over\newline 2012–01 to 2016–12\newline (monthly)
& Produces non-informative diagnosis timing for diabetes, CKD, and AF, emphasising that EMR diagnosis timestamps often do not align with biological onset or event timing. \\\\

\hline
\end{tabular}
\label{tab:messiness_sampling}
\end{table}

%===========================================================
\newpage
\begin{table}[htbp]
\footnotesize
\centering
\caption{Injected Irregularities in \texttt{[PatientEMR].[MasterSummary]}}
\begin{tabular}{p{3cm} p{3cm} p{6cm}}
\hline
\textbf{Location} & \textbf{Irregularity} & \textbf{Notes} \\
\hline

\texttt{Patient\_ID} &
Non-sequential\newline identifier &
-- Computed as $(i^2 - 77)\times 3 + 500$ to obscure \newline\hspace*{5mm}ordering.\newline
-- Mirrors de-identified EMR IDs with no intuitive \newline\hspace*{5mm}structure.\newline
-- Keeps deterministic linkage across tables. \\\\[4pt]

\texttt{Age\_At\_2024} &
Shifted age reference year &
-- Age stored as ``Age at 2024'', computed as \newline\hspace*{5mm}Age + 7.\newline
-- Reflects EMRs where age corresponds to \newline\hspace*{5mm}extraction year, not event time.\newline
-- Adds mild temporal misalignment relative to \newline\hspace*{5mm}baseline. \\\\[4pt]

\texttt{SMOKING\_STATUS} &
Patterned\newline missingness &
-- Starts from true smoking status in clean cohort.\newline
-- 15.66\% of non-smokers set to \texttt{NaN}.\\\\[4pt]

\texttt{IRSD\_Quintile} &
Socioeconomic index unchanged &
-- Passed through directly from clean dataset.\newline
-- Reflects stable area-level deprivation coding.\newline
-- No measurement noise introduced here. \\\\[4pt]

\texttt{CVD\_Event} &
Integer event flag &
-- Converted to integer 0/1 representation.\newline
-- Matches common EMR export conventions.\newline
-- Supports downstream survival analysis \newline\hspace*{5mm}pipelines. \\\\[4pt]

\texttt{CVD\_Time} &
Coarsened calendar event time &
-- True follow-up time mapped to calendar date \newline \hspace*{5mm}via 2017--01--01.\newline
-- Converted to \texttt{YYYY-MM} format (month-level \newline \hspace*{5mm}resolution).\newline
-- Discards daily granularity typical of EMR \newline \hspace*{5mm}codification. \\\\[4pt]

\texttt{CVD\_Time} &
Common\newline administrative\newline censor date &
-- All non-events assigned \texttt{"2022-12"}.\newline
-- Produces realistic right-censoring uniformity. \\\\[4pt]

\hline
\end{tabular}
\end{table}

%===========================================================
\newpage
\begin{table}[htbp]
\footnotesize
\centering
\caption{Injected Irregularities in \texttt{[PatientEMR].[ChronicDiseases]}}
\begin{tabular}{p{3cm} p{3cm} p{6cm}}
\hline
\textbf{Location} & \textbf{Irregularity} & \textbf{Notes} \\
\hline
\texttt{Patient\_ID} &
Presence-only\newline condition table &
-- Contains rows only for patients with \newline \hspace*{5mm}diabetes, CKD, or AF.\newline
-- Same nonlinear ID transformation used as in \newline \hspace*{5mm}other tables. \\\\[4pt]

\texttt{Category} (Diabetes) &
Diabetes\newline label variability &
-- Canonical ``Diabetes'' replaced by multinomial \newline \hspace*{5mm}labels.\newline
-- Includes ICD codes, abbreviations, and lay \newline \hspace*{5mm}labels.\newline
-- Introduces semantic variability requiring \newline \hspace*{5mm}harmonisation. \\\\[4pt]

\texttt{Category} (AF) &
AF\newline label variability &
-- Similar randomness to diabetes labels. \\\\[4pt]

\texttt{Category} (CKD) &
CKD\newline label variability &
-- Similar randomness to diabetes labels. \\\\[4pt]

\texttt{Date} &
Randomised\newline diagnosis dates &
-- Date assigned uniformly from 2012--01 \newline \hspace*{5mm}to 2016--12.\newline
-- Not linked to biological onset or CVD event \newline \hspace*{5mm}time.\newline
-- Rows permuted using fixed RNG.\newline
-- Prevents inference of sorting or grouping \newline \hspace*{5mm}structure.\\\\[4pt]

\hline
\end{tabular}
\end{table}

%===========================================================
\newpage
\begin{table}[htbp]
\footnotesize
\centering
\caption{Injected Irregularities in \texttt{[PatientEMR].[MeasAndPath]}}
\begin{tabular}{p{3cm} p{3cm} p{6cm}}
\hline
\textbf{Location} & \textbf{Irregularity} & \textbf{Notes} \\
\hline

\texttt{Patient\_ID} &
 &
-- Same nonlinear ID transformation used as in \newline \hspace*{5mm}other tables.\\\\[4pt]

\texttt{Measure} &
Stacked long-form\newline biomarkers &
-- HbA1c, eGFR, SBP concatenated into one \newline \hspace*{5mm}table.\newline
-- Requires pivoting or filtering to build analytic \newline \hspace*{5mm}datasets.\\\\[4pt]

\texttt{Description} (HbA1c) &
Heterogeneous labels &
-- Includes ``HbA1c'', ``HBA1C'', ``A1C'',\newline \hspace*{5mm}``Glycated hemoglobin'', etc.\newline
-- Probability-weighted distribution.\\\\[4pt]

\texttt{Description} (eGFR) &
Heterogeneous labels &
-- Includes ``eGFR'', ``EGFR'', ``GFR'', \newline \hspace*{5mm}``Estimated GFR'', and variants.\newline
-- Probability-weighted distribution.\\\\[4pt]

\texttt{Description} (SBP) &
Heterogeneous labels &
-- Includes ``SBP'', ``Systolic BP'', ``BP Systolic'', \newline \hspace*{5mm} and variants.\newline
-- Probability-weighted distribution.\\\\[4pt]

\texttt{Unit} (canonical) &
Assigned\newline deterministically &
-- SBP → mmHg.\newline
-- eGFR → mL/min/1.73m$^2$.\newline
-- HbA1c → \%. \\\\[4pt]

\texttt{Value/Unit} (HbA1c converted) &
Mixed-unit\newline representation &
-- 5\% of HbA1c values converted to mmol/mol.\newline
-- Uses linear transformation $(x - 2.15)\times 10.929$.\\\\[4pt]

\texttt{Date} &
Randomised test\newline timing &
-- Month sampled uniformly from \newline \hspace*{5mm}2012--01 to 2016--12.\newline
-- Independent of cohort baseline timing.\newline
-- Rows permuted using fixed RNG.\newline
-- Removes structure in ordering by patient \newline \hspace*{5mm}or measure.\\\\[4pt]

\hline
\end{tabular}
\end{table}

%===========================================================
%===========================================================
%===========================================================
\newpage
\section{Student Questions in Technical Validation}

\subsection{Q1: Exploratory Reconstruction and Socioeconomic Comparison (10 marks)}~\label{App:Val10}

You are provided with Data Asset 2, a relational, EMR-style dataset consisting of three linked tables:\\
\hspace*{5mm}(1) \texttt{PatientMasterSummary}\\
\hspace*{5mm}(2) \texttt{PatientChronicDiseases}\\
\hspace*{5mm}(3) \texttt{PatientMeasAndPath}

These tables contain patient records with heterogeneous diagnosis labels, non-sequential identifiers, and structured demographic information.

\textbf{Task}\\
Using Data Asset 2 only, complete the following tasks:\\
\hspace*{ 5mm}(a) Cohort reconstruction (CKD vs T2DM).\\
\hspace*{10mm}Reconstruct two mutually exclusive patient cohorts:\\
\hspace*{15mm}(i) patients with a recorded diagnosis of chronic kidney disease (CKD) only, and\\
\hspace*{15mm}(ii) patients with a recorded diagnosis of type~2 diabetes mellitus (T2DM) only.\\
\hspace*{10mm}Diagnoses must be identified from the \texttt{PatientChronicDiseases} table, appropriately\\
\hspace*{10mm}handling heterogeneous free-text and code-like labels. Patients must be linked across tables\\
\hspace*{10mm}using the provided patient identifiers.

\hspace*{ 5mm}(b) Socioeconomic summarisation.\\
\hspace*{10mm}For each reconstructed cohort, compute the prevalence (percentage) of patients in each Index\\
\hspace*{10mm}of Relative Socioeconomic Disadvantage (IRSD) quintile (1--5), using the IRSD information\\
\hspace*{10mm}stored in \texttt{PatientMasterSummary}.

\hspace*{ 5mm}(c) Visualisation.\\
\hspace*{10mm}Produce a side-by-side bar plot comparing the IRSD quintile distributions of the CKD-only\\
\hspace*{10mm}and T2DM-only cohorts, with the following specifications:\\
\hspace*{15mm}(i) y-axis: prevalence (\%);\\
\hspace*{15mm}(ii) x-axis: IRSD quintiles;\\
\hspace*{15mm}(iii) colour: T2DM-only cohort in cyan and CKD-only cohort in magenta;\\
\hspace*{15mm}(iv) appropriate axis labels and a legend must be included.

\hspace*{ 5mm}(d) Interpretation.\\
\hspace*{10mm}In 2--3 sentences, briefly describe the key differences or similarities observed between the two\\
\hspace*{10mm}cohorts’ socioeconomic profiles.

A correct answer will demonstrate successful linkage of relational EMR tables, accurate reconstruction of mutually exclusive disease cohorts, correct computation of IRSD prevalence, appropriate visualisation choices, and a concise, data-driven interpretation consistent with the displayed results.

\subsection{Q1 Suggested Solution}~\label{App:Val11}
\begin{lstlisting}[language=Python, style=mystyle, backgroundcolor=\color{backcolour4Input}]
import os, re
import numpy as np
import pandas as pd
import matplotlib.pyplot as plt

def seed_all(seed=42):
    np.random.seed(seed)

seed_all()

BASE_PATH = "" # User's drive
PAT_FILE  = os.path.join(BASE_PATH, 
                         "Data002_PatientEMR_MasterSummary.csv")
CHRO_FILE = os.path.join(BASE_PATH, 
                         "Data002_PatientEMR_ChronicDiseases.csv")

def safe_read_csv(path):
    df = pd.read_csv(path)
    return df.loc[:, ~df.columns.str.contains("^Unnamed")]

pat = safe_read_csv(PAT_FILE)
chro = safe_read_csv(CHRO_FILE)

diag_col_candidates = \
    ["Category", "category",
     "Classification", "classification",
     "Label", "label",
     "Classification ", "Category "]

diag_col = next((c for c in chro.columns if c in diag_col_candidates), None)
if diag_col is None:
    text_cols = [c for c in chro.columns if chro[c].dtype == object]
    diag_col = text_cols[0]

chro['diag_text'] = chro[diag_col].astype(str).str.lower()

CKD_PATTERNS  = [r"\bckd\b", r"\bchronic kidney\b", 
                 r"\bchronic renal\b", r"\brenal insuf", 
                 r"\brenal fail", r"\bkidney failure\b", r"\begfr\b"]
T2DM_PATTERNS = [r"\bt2dm\b", r"\btype ?2\b", r"\btype 2\b", 
                 r"\btype ?ii\b", r"\bdiabetes\b", r"\be11\b"]

def matches_any(text, patterns):
    for p in patterns:
        if re.search(p, text):
            return True
    return False

chro['is_ckd_record']  = \
    chro['diag_text'].apply(lambda x: matches_any(x, CKD_PATTERNS))
chro['is_t2dm_record'] = \
    chro['diag_text'].apply(lambda x: matches_any(x, T2DM_PATTERNS))

flags = chro.groupby('Patient_ID').\
        agg({'is_ckd_record':'any','is_t2dm_record':'any'}).\
        reset_index().\
        rename(columns={'is_ckd_record':'has_ckd',
                        'is_t2dm_record':'has_t2dm'})

pat = pat.merge(flags, on='Patient_ID', how='left')
pat['has_ckd']  = pat['has_ckd'].fillna(False)
pat['has_t2dm'] = pat['has_t2dm'].fillna(False)
pat['CKD_only']  = pat['has_ckd'] & (~pat['has_t2dm'])
pat['T2DM_only'] = pat['has_t2dm'] & (~pat['has_ckd'])

irsd_candidates = ['IRSD_Quintile', 'IRSD', 'irsd_quintile', 
                   'irsd', 'IRSD Quintile', 'IRSD_Quintile']
irsd_col = next((c for c in pat.columns if \
                 c in irsd_candidates), None)
if irsd_col is None:
    num_cols = [c for c in pat.columns if \ 
                np.issubdtype(pat[c].dropna().dtype, np.number)]
    found = None
    for c in num_cols:
        vals = pd.Series(pat[c].dropna().unique()).astype(int)
        if set(vals).issubset({1,2,3,4,5}):
            found = c
            break
    irsd_col = found

pat[irsd_col] = pd.to_numeric(pat[irsd_col], errors='coerce').\
                round().astype('Int64')

def prevalence(df, mask, col):
    s = df.loc[mask, col].dropna().astype(int)
    total = len(s)
    counts = s.value_counts(sort=False).\
             reindex([1,2,3,4,5], fill_value=0)
    pct = (counts / total * 100.0) if total>0 else (counts*0.0)
    return pct.values

ckd_pct  = prevalence(pat, pat['CKD_only'],  irsd_col)
t2dm_pct = prevalence(pat, pat['T2DM_only'], irsd_col)

x_base = np.array([1,3,5,7,9])
half = 0.18
x_ckd  = x_base - half
x_t2dm = x_base + half

fig, ax = plt.subplots(figsize=(9,5))
ax.bar(x_t2dm, t2dm_pct, width=0.36, label='T2DM only', 
       color='cyan',    alpha=0.5, edgecolor='k', linewidth=0.4)
ax.bar(x_ckd,  ckd_pct,  width=0.36, label='CKD only',  
       color='magenta', alpha=0.5, edgecolor='k', linewidth=0.4)
ax.set_xticks(x_base)
ax.set_xticklabels(['1','2','3','4','5'])
ax.set_xlim(0,10)
ax.set_ylim(0,50)
ax.set_ylabel('Prevalence (%)')
ax.set_xlabel('IRSD quintile')
ax.set_title(
    'IRSD distribution - CKD only vs T2DM only '+\
    '(prevalence % by quintile)')
ax.legend(frameon=True)
ax.yaxis.grid(True, linestyle='--', linewidth=0.5, alpha=0.6)
plt.tight_layout()
plt.show()
plt.close(fig)

\end{lstlisting}

\subsection{Q2: Socioeconomic Stratification and Distributional Assessment (10 marks)}~\label{App:Val20}

You are provided with \textbf{Data Asset 1}, a clean, analysis-ready cohort containing one row per simulated individual with demographic, socioeconomic, behavioural, clinical, and cardiovascular outcome variables.

\textbf{Task}\\
Using Data Asset 1 only, complete the following tasks:\\
\hspace*{5mm}(a) Stratification by socioeconomic status.\\
\hspace*{10mm}Stratify the cohort by Index of Relative Socioeconomic Disadvantage (IRSD) quintile (1--5).\\
\hspace*{5mm}(b) Distributional summarisation.\\
\hspace*{10mm}For each IRSD quintile, summarise the distribution of the following variables:\\
\hspace*{15mm}age, smoking status, body mass index (BMI), systolic blood pressure (SBP),\\
\hspace*{15mm}HbA1c, estimated glomerular filtration rate (eGFR), diabetes,\\
\hspace*{15mm}chronic kidney disease (CKD), atrial fibrillation (AF), and\\
\hspace*{15mm}5-year cardiovascular disease (CVD) outcome.\\
\hspace*{5mm}(c) Visualisation.\\
\hspace*{10mm}Produce appropriate IRSD-stratified visualisations, including:\\
\hspace*{15mm}(i) boxplots for continuous variables;\\
\hspace*{15mm}(ii) stacked bar or count plots for categorical and binary variables.\\
\hspace*{10mm}All figures must include clearly labelled axes and legends.\\
\hspace*{5mm}(d) Interpretation and modelling implications.\\
\hspace*{10mm}In 3--4 sentences, describe the key socioeconomic gradients observed.

A correct answer will demonstrate appropriate stratification, coherent distributional summaries, effective visualisation choices, and a concise interpretation linking socioeconomic structure to calibration and fairness considerations in downstream risk modelling.

%\newpage
\subsection{Q2 Suggested Solution}~\label{App:Val21}
\begin{figure}[h!]
    \centering
    \includegraphics[width=\linewidth]{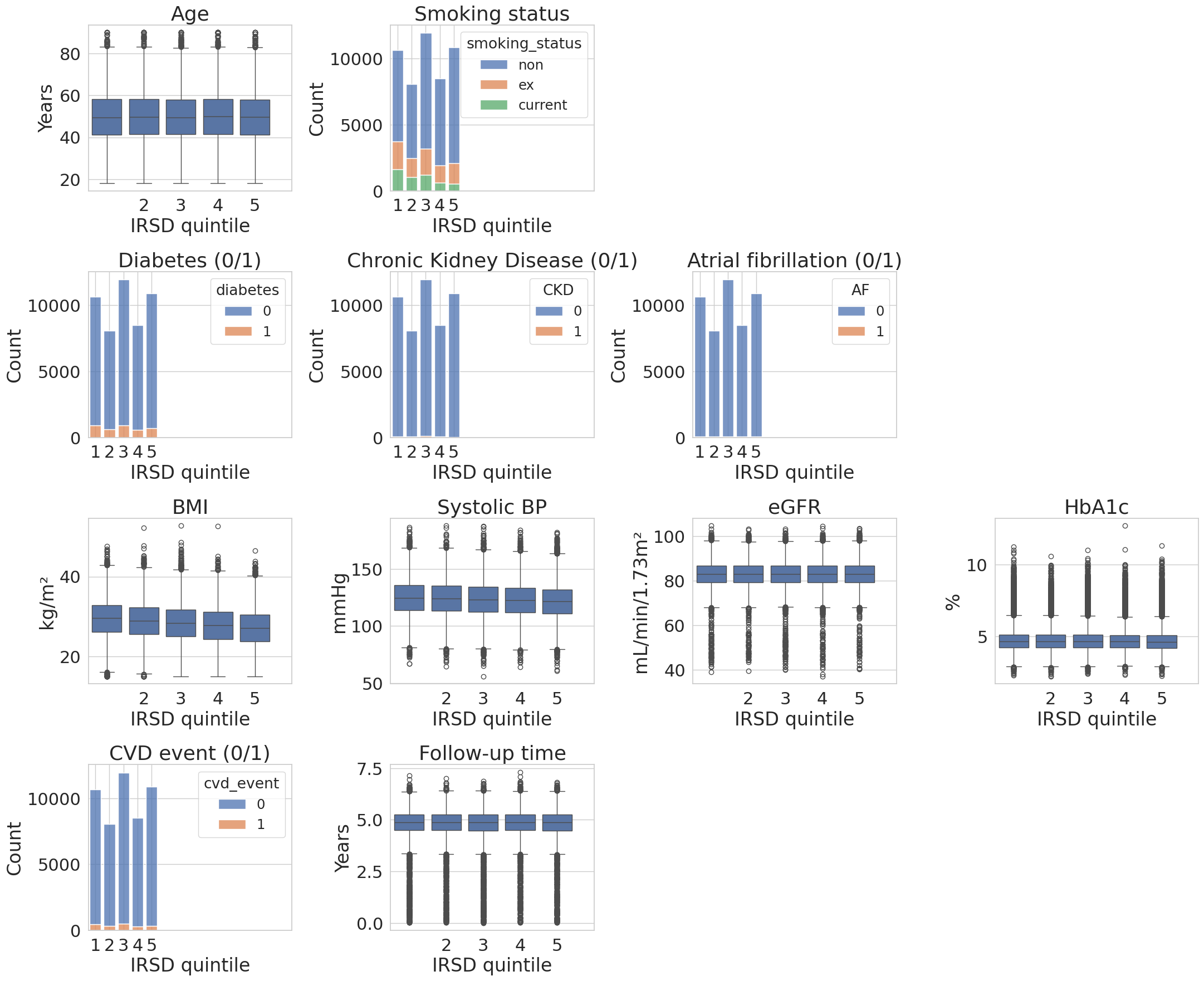}
    \caption{IRSD-stratified distributions of variables in the PRIME-CVD cohort.}
\end{figure}
This figure summarises how key demographic, clinical, lifestyle, and cardiovascular outcome variables vary across IRSD quintiles within the PRIME-CVD cohort. For numeric variables, side-by-side boxplots reveal modest socioeconomic gradients -- for example, BMI, SBP, and HbA1c show small but progressive decreases with increasing IRSD (\textit{i.e.,} decreasing disadvantage), while eGFR and follow-up time remain broadly consistent across quintiles. Stacked histograms for categorical variables illustrate expected socioeconomic patterns in health behaviours and chronic disease, including higher proportions of current smokers and diabetes cases in more disadvantaged groups. Conversely, conditions with low prevalence (CKD, AF) exhibit only minimal variation across IRSD.

\subsection{Q3: Multivariable Hazard Modelling and Policy Interpretation (10 marks)}~\label{App:Val30}

You are provided with Data Asset 1, a clean, analysis-ready cohort containing one row per simulated individual with demographic, socioeconomic, behavioural, clinical, and cardiovascular outcome variables, including time-to-event and censoring information.

\textbf{Task}\\
Using Data Asset 1 only, complete the following tasks:\\
\hspace*{5mm}(a) Model specification and fitting.\\
\hspace*{10mm}Fit a multivariable Cox proportional hazards model to estimate 5-year CVD risk.\\
\hspace*{10mm}Use non-smokers and IRSD quintile~5 as reference categories.\\
\hspace*{5mm}(b) Estimation and summarisation.\\
\hspace*{10mm}Report adjusted hazard ratios with 95\% confidence intervals and \textit{p}-values\\
\hspace*{10mm}for all model covariates in a clearly formatted table.\\
\hspace*{5mm}(c) Visualisation.\\
\hspace*{10mm}Produce a forest plot displaying hazard ratios and their 95\% confidence intervals,\\
\hspace*{10mm}with covariates on the y-axis and a vertical reference line at hazard ratio~=~1.

A correct answer will demonstrate appropriate model specification, numerically stable estimation, clear presentation of adjusted hazard ratios, effective visualisation of uncertainty, and a concise interpretation linking multivariable survival modelling to equitable cardiovascular risk assessment.

\subsection{Q3 Suggested Solution}~\label{App:Val31}
\begin{lstlisting}[language=Python, style=mystyle, backgroundcolor=\color{backcolour4Input}]
import numpy as np
import pandas as pd
import matplotlib.pyplot as plt
from lifelines import CoxPHFitter

seed_all()

df = Data001.copy()
df["Age_c"] = df["Age"] - 30.0

irsd_dummies = pd.get_dummies(df["IRSD_quintile"], prefix="irsd",
                              drop_first=False)
if "irsd_5" in irsd_dummies.columns:
    irsd_dummies = irsd_dummies.drop(columns=["irsd_5"])

smoke_dummies = pd.get_dummies(df["smoking_status"], prefix="smoke",
                               drop_first=False)
if "smoke_non" in smoke_dummies.columns:
    smoke_dummies = smoke_dummies.drop(columns=["smoke_non"])

for col in ["AF", "CKD", "diabetes"]:
    df[col] = df[col].astype(int)

X = pd.concat([df[["Age_c", 
                    "AF", "CKD", "diabetes", 
                    "HbA1c", "BMI", "eGFR", "SBP"]], 
                    smoke_dummies, irsd_dummies], axis=1)

events = df["cvd_event"].astype(bool)
cols_to_drop = []
for col in X.columns:
    v_all = X[col].var()
    v_event = X.loc[events, col].var() if \
                events.sum() > 0 else \
                0.0
    v_nonevent = X.loc[~events, col].var() if 
                (~events).sum() > 0 else \
                0.0
    if (v_all < 1e-6) or (v_event < 1e-6) or (v_nonevent < 1e-6):
        cols_to_drop.append(col)

X_reduced = X.drop(columns=cols_to_drop)
cox_df = pd.concat([df[["cvd_time", "cvd_event"]], X_reduced], axis=1)

cph = CoxPHFitter(penalizer=0.05)
cph.fit(cox_df, 
        duration_col="cvd_time", event_col="cvd_event", 
        show_progress=False)

hr_table = cph.summary[["exp(coef)", 
                        "exp(coef) lower 95%", "exp(coef) upper 95%", 
                        "p"]].\
                        rename(columns={
                                "exp(coef)": "HR",
                                "exp(coef) lower 95%": "HR_lower_95",
                                "exp(coef) upper 95%": "HR_upper_95",
                                "p": "p_value",
                                })

def rename_hr_variables(hr_table):
    rename_map = {"Age_c": "Age", 
                  "smoke_current": "smoke (cur)", 
                  "smoke_ex": "smoke (ex)"}
    renamed_index = []
    for var in hr_table.index:
        if var in rename_map:
            renamed_index.append(rename_map[var])
        elif var.startswith("irsd_"):
            q = var.split("_")[1]
            renamed_index.append(f"IRSD: {q}")
        else:
            renamed_index.append(var)
    hr_table_renamed = hr_table.copy()
    hr_table_renamed.index = renamed_index
    return hr_table_renamed

hr_table_plot = rename_hr_variables(hr_table)

def plot_forest_hr(hr_table, title=None):
    df = hr_table.copy()
    variables = df.index.tolist()
    y_pos = np.arange(len(variables))
    hr = df["HR"].values
    lower = df["HR_lower_95"].values
    upper = df["HR_upper_95"].values
    xerr = np.vstack([hr - lower, upper - hr])
    fig, ax = plt.subplots(figsize=(8, 0.45 * len(variables)))
    ax.errorbar(hr, y_pos, xerr=xerr, fmt="o", 
                color="black", ecolor="black", 
                elinewidth=1.5, capsize=3)
    ax.axvline(x=1.0, linestyle="--", color="gray", linewidth=1.2)
    ax.set_yticks(y_pos)
    ax.set_yticklabels(variables)
    ax.set_xlabel("Hazard Ratio (HR)")
    ax.set_ylabel("Covariate")
    ax.invert_yaxis()
    ax.grid(axis="x", linestyle=":", alpha=0.6)
    if title:
        ax.set_title(title)
    plt.tight_layout()
    plt.show()
    plt.close(fig)

plot_forest_hr(hr_table_plot, 
               title="Cox proportional hazards model - Data Asset 1")


\end{lstlisting}

%%%###===###%%%###===###%%%###===###%%%###===###%%%###===###%%%###===###
%%%###===###%%%###===###%%%###===###%%%###===###%%%###===###%%%###===###
%%%###===###%%%###===###%%%###===###%%%###===###%%%###===###%%%###===###
\newpage
\section{Additional Validations}
\subsection{A Complete Epidemiologic Correlation of the PRIME-CVD Cohort}
\begin{figure}[h!]
    \centering
    \includegraphics[width=\linewidth]{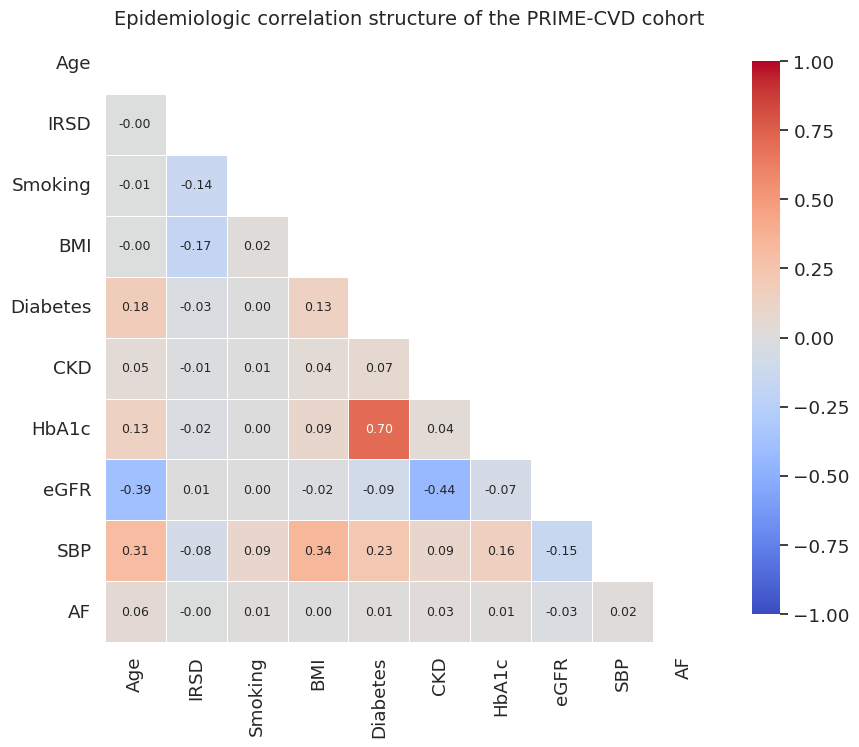}
    \caption{Lower-triangular Pearson correlation matrix for key variables in the PRIME-CVD cohort.}
\end{figure}

This correlation matrix summarises the epidemiologic associations between socioeconomic, behavioural, anthropometric, and physiological variables in the PRIME-CVD cohort. Smoking status was temporarily encoded as an ordinal variable (“non”<“ex”<“current”) solely to allow its inclusion in a numerical Pearson framework. The overall pattern shows expected relationships -- such as the strong association between diabetes and HbA1c, moderate positive correlations between systolic blood pressure and age or BMI, and moderate negative correlations between eGFR and age. At the same time, many other pairwise correlations appear close to zero. This is an anticipated property of the PRIME-CVD design: the dataset is generated mechanistically from a hand-specified causal DAG derived from high-level AIHW/ABS statistics rather than from patient-level electronic records or machine-learned generative models. As such, only relationships explicitly encoded in the DAG manifest as correlations, while unmodelled or subtle dependencies remain absent. For this reason, this figure is presented in the appendix: it provides transparency about the underlying epidemiologic structure, while reinforcing that PRIME-CVD reflects a targeted, pedagogically oriented simulation rather than a comprehensive representation of all real-world clinical correlations.

\newpage
\subsection{Stratification Along Age Groups}
\begin{figure}[h!]
    \centering
    \includegraphics[width=\linewidth]{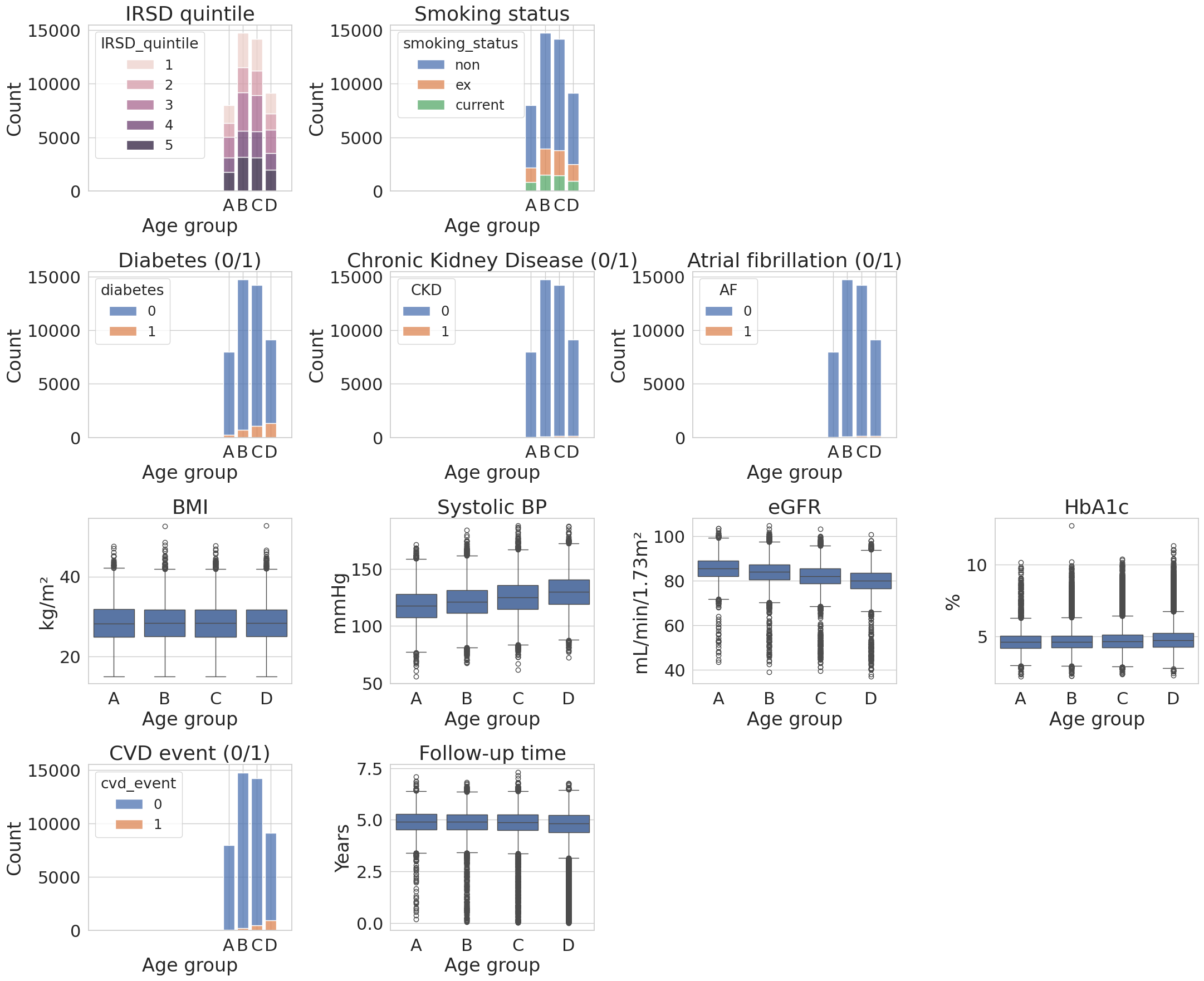}
    \caption{Age-stratified distributions of variables in the PRIME-CVD cohort.}
\end{figure}

This figure presents the distribution of key socioeconomic, behavioural, clinical, and outcome variables across four age groups (A: 30–39, B: 40–49, C: 50–59, D: 60–74) within the PRIME-CVD cohort. Stacked barplots show clear age-associated patterns in categorical features, including increasing prevalence of diabetes, CKD, AF, and CVD events with advancing age, alongside a shift toward higher IRSD variability in older strata. Smoking patterns also evolve modestly, with current smoking more common in younger adults and ex-smoking more common in older adults. Boxplots for anthropometric and physiological measurements illustrate expected age-related gradients: SBP and HbA1c rise progressively with age, while eGFR declines, reflecting the intended physiologic realism embedded in the model. Conversely, BMI shows only a mild age-related shift, consistent with the comparatively weaker age–BMI link specified in the underlying causal DAG.

\newpage
\begin{table}[h!]
\centering
\scriptsize
\renewcommand{\arraystretch}{1.25}
\begin{tabular}{p{0.50cm} 
                p{0.70cm} 
                p{1.80cm} 
                p{0.70cm} 
                p{0.50cm} 
                p{0.50cm} 
                p{1.15cm} 
                p{1.15cm} 
                p{1.15cm} 
                p{1.15cm} 
                p{0.70cm} }
\hline
\textbf{Age} & 
\textbf{N} & 
\textbf{IRSD\newline (\%)} & 
\textbf{Diabetes\newline (\%)} & 
\textbf{CKD\newline (\%)} & 
\textbf{AF\newline (\%)} &
\textbf{BMI\newline mean (SD)} &
\textbf{SBP\newline mean (SD)} &
\textbf{eGFR\newline mean (SD)} &
\textbf{HbA1c\newline mean (SD)} &
\textbf{Event\newline (\%)} \\
\hline

N & 
2{,}878 & 
1: 22.03\newline2: 15.71\newline3: 23.00\newline4: 17.23\newline5: 22.03 & 
1.74 & 0.28 & 0.10 & 
28.28\newline(4.97) & 
113.41\newline(14.86) & 
87.39\newline(5.28) & 
4.62\newline(0.70) & 
0.21 \\\\

A & 
7{,}976 & 
1: 20.84\newline2: 16.12\newline3: 23.92\newline4: 16.76\newline5: 22.35 & 
2.95 & 0.29 & 0.18 & 
28.34\newline(5.11) & 
117.82\newline(15.24) & 
85.31\newline(5.33) & 
4.67\newline(0.75) & 
0.56 \\\\

B & 
14{,}746 & 
1: 21.66\newline2: 16.00\newline3: 24.40\newline4: 16.64\newline5: 21.31 & 
4.90 & 0.41 & 0.49 & 
28.36\newline(5.02) & 
121.42\newline(15.17) & 
83.71\newline(5.41) & 
4.72\newline(0.82) & 
1.49 \\\\

C & 
14{,}202 & 
1: 21.09\newline2: 16.13\newline3: 23.55\newline4: 17.19\newline5: 22.04 & 
7.72 & 0.80 & 0.82 & 
28.31\newline(5.02) & 
125.26\newline(15.48) & 
81.83\newline(5.74) & 
4.79\newline(0.94) & 
3.46 \\\\

D & 
9{,}121 & 
1: 21.16\newline2: 16.32\newline3: 23.84\newline4: 17.19\newline5: 21.49 & 
14.54 & 1.26 & 1.44 & 
28.35\newline(5.00) & 
129.87\newline(15.63) & 
79.70\newline(6.13) & 
4.97\newline(1.16) & 
10.28 \\\\

K & 
1{,}077 & 
1: 20.89\newline2: 16.53\newline3: 23.68\newline4: 18.48\newline5: 20.43 & 
26.74 & 1.86 & 2.23 & 
28.09\newline(5.05) & 
135.24\newline(16.17) & 
77.07\newline(6.61) & 
5.30\newline(1.39) & 
28.69 \\\\

\hline
\end{tabular}
\caption{Characteristics of the PRIME-CVD cohort stratified by age group (N: 18--29, A: 30--39, B: 40--49, C: 50--59, D: 60--74, K: $\geq$75).}
\label{tab:prime_age_irsd}
\end{table}

Examination of the youngest (N: 18–29) and oldest (K: $\ge$75) age groups in the PRIME-CVD cohort highlights both expected epidemiologic contrasts and the intended structural boundaries of a DAG-driven simulation. Group N displays uniformly low prevalences of diabetes, CKD, AF, and CVD events, alongside higher eGFR and lower SBP, reflecting physiologic advantage in early adulthood. In contrast, Group K shows markedly elevated rates of chronic disease and CVD events, higher SBP and HbA1c, and noticeably reduced kidney function. Although these gradients align with real-world aging trajectories, other patterns -- such as the relatively flat IRSD distribution across age bands -- reflect the fact that age–IRSD relationships were not encoded in the underlying DAG, producing a limitation analogous to that seen in the correlation matrix, where variables without explicit causal links exhibit near-zero associations. This behaviour is expected for a parametrically defined model built from high-level AIHW and ABS summary statistics rather than patient-level data, and it reinforces that PRIME-CVD is designed as a pedagogically oriented environment rather than a fully comprehensive epidemiologic reconstruction. Nonetheless, PRIME captures the principal cardiometabolic trends needed for teaching, allowing learners to engage with realistic age-related disease patterns.

\newpage
\subsection{Cleaning and Linking the Relational Data Asset}
If the relational EMR-style asset were used as the sole input for cohort construction, the first step would involve deriving a cleaned patient-level table from \texttt{[PatientEMR].[MasterSummary]}. Age is reconstructed by subtracting seven years from \texttt{Age\_At\_2024}, socioeconomic position and smoking status are restored through simple renaming, and the coarse year--month CVD outcome field is mapped back to a continuous follow-up time by anchoring at 1~January~2017 and assigning the 15th day of each recorded month as the event or censoring date. This produces a baseline structure containing \texttt{Age}, \texttt{IRSD\_quintile}, \texttt{smoking\_status}, \texttt{cvd\_event} and \texttt{cvd\_time}, mirroring the static structure of PRIME-CVD while relying solely on information that would plausibly be present in an EMR extract.

The second step requires recovering chronic disease indicators from\\ \texttt{[PatientEMR].[PatientChronicDiseases]}. We define finite vocabularies of free-text and coded terms corresponding to diabetes, CKD and AF, and map each heterogeneous \texttt{Category} entry into a canonical condition label. After discarding rows that do not match these dictionaries, we collapse the long-format table to a wide one with one row per patient and indicator columns (\texttt{diabetes}, \texttt{CKD}, \texttt{AF}) obtained by taking the maximum across all diagnosis rows for a given patient. Left-joining these reconstructed indicators back to the patient-level table, with missing values imputed to zero, recovers the essential chronic disease structure present in the original, fully harmonised PRIME-CVD dataset.

Finally, biomarker harmonisation is performed using \texttt{[PatientEMR].[PatientMeasAndPath]}. We subset the long-form table into HbA1c, eGFR and SBP measurements, resolve lexical variability in the test descriptions, and correct unit inconsistencies by converting any HbA1c values recorded in mmol/mol back to percentage units using the inverse of the forward transformation. For each biomarker, we retain a single canonical numeric value per patient and merge these onto the reconstructed cohort via \texttt{Patient\_ID}. The resulting dataset -- containing \texttt{Age}, \texttt{IRSD\_quintile}, \texttt{smoking\_status}, \texttt{diabetes}, \texttt{CKD}, \texttt{AF}, \texttt{HbA1c}, \texttt{eGFR}, \texttt{SBP}, \texttt{cvd\_event} and \texttt{cvd\_time} -- approximates the original PRIME-CVD static asset, acknowledging that some information (\textit{e.g.,} BMI, exact event timing, multi-episode structure) is irretrievably lost during EMR decomposition. Nonetheless, this reconstruction process faithfully illustrates the data cleaning, harmonisation and linkage challenges that arise in transforming real-world relational EMR data into analysis-ready cohorts.

\newpage
\begin{figure}[h!]
    \centering
    \includegraphics[width=1.15\linewidth]{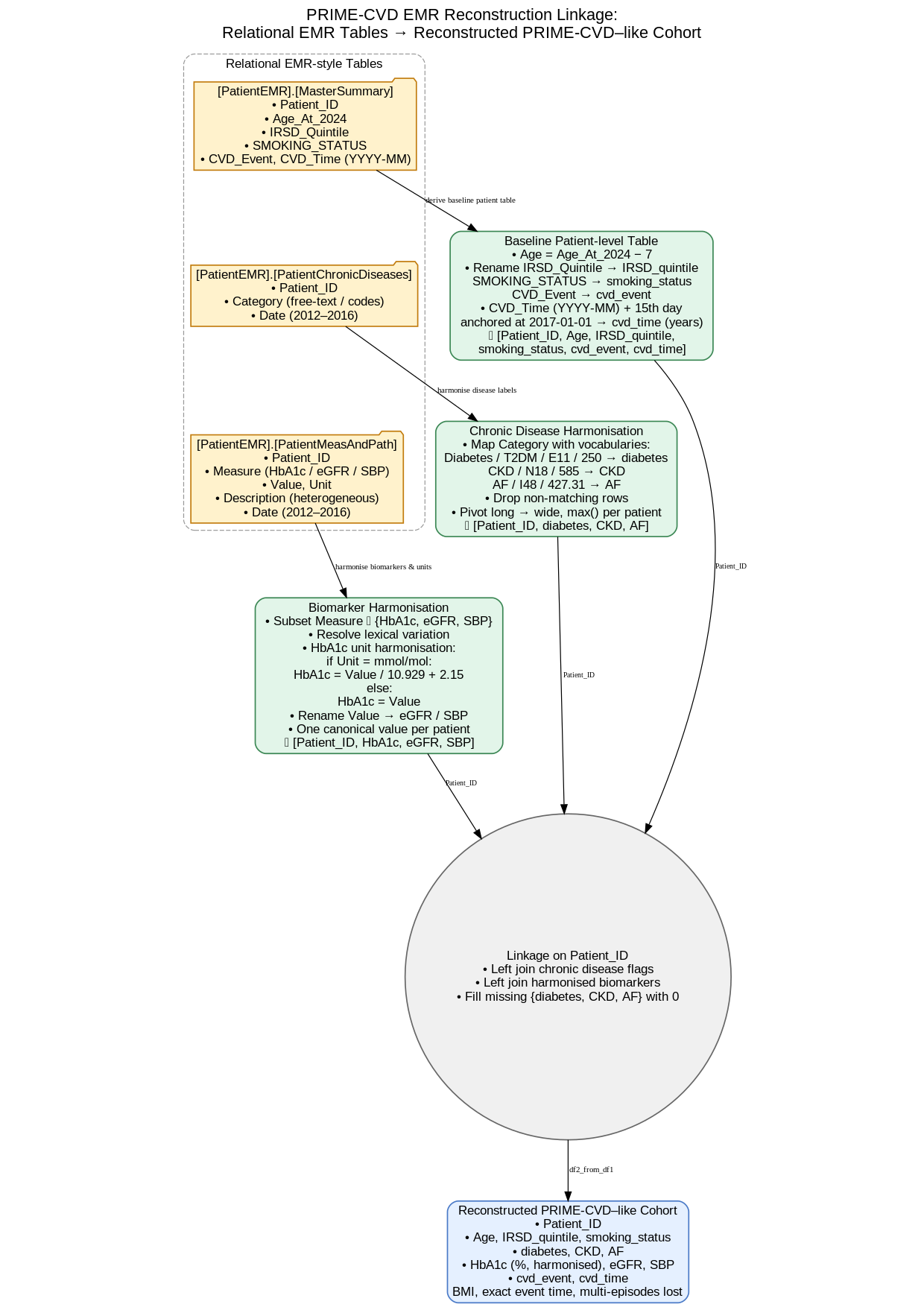}
    \caption{Suggested Reconstruction Pipeline}
\end{figure}

\end{document}